\documentclass{article}

\usepackage[preprint,nonatbib]{neurips_2021}

\usepackage[utf8]{inputenc} % allow utf-8 input
\usepackage[T1]{fontenc}    % use 8-bit T1 fonts
\usepackage{url}            % simple URL typesetting
\usepackage{booktabs}       % professional-quality tables
\usepackage{amsfonts}       % blackboard math symbols
\usepackage{subfigure}
\usepackage{nicefrac}       % compact symbols for 1/2, etc.
\usepackage{microtype}      % microtypography
\usepackage{xcolor}         % colors
\usepackage{amsmath}
\usepackage{wrapfig}
\usepackage{graphicx}
\usepackage{amssymb}
\usepackage{algorithm}
\usepackage{algorithmic}
\usepackage{multirow}
\usepackage{diagbox}
\usepackage{booktabs}
\usepackage{ntheorem}
\usepackage{bm}
\usepackage{colortbl}
\usepackage[colorlinks,
            linkcolor=red,
            anchorcolor=blue,
            citecolor=brown,
            ]{hyperref}
\usepackage{subfigure}
\usepackage{parskip}
\usepackage{stmaryrd}
\usepackage{caption}
\usepackage{varwidth}

\newtheorem{definition}{Definition}
\newtheorem{theorem}{Theorem}
\newtheorem{lemma}{Lemma}
\newtheorem*{proof}{Proof.}[section]

\newcommand{\bb}{\hspace{-1mm} $\bullet$}

\title{On Understanding and Mitigating the Dimensional Collapse of Graph Contrastive Learning: a Non-Maximum Removal Approach}

\author{Jiawei Sun, Ruoxin Chen, Jie Li, Chentao Wu, Yue Ding, Junchi Yan\\
Shanghai Jiao Tong University\\
\texttt{\{noelsjw, chenruoxin, lijiecs, wuct, dingyue, yanjunchi\}@sjtu.edu.cn}\\
}

\begin{document}

\maketitle
\begin{abstract}
Graph Contrastive Learning (GCL) has shown promising performance in graph representation learning (GRL) without the supervision of manual annotations. 
GCL can generate graph-level embeddings by maximizing the Mutual Information (MI) between different augmented views of the same graph (positive pairs). 
However, the GCL is limited by dimensional collapse, i.e., embedding vectors only occupy a low-dimensional subspace.  
In this paper, we show that the smoothing effect of the permutation-invariant graph pooling and the implicit regularization of the graph convolution are two 
causes of the dimensional collapse in GCL.
To mitigate the above issue, we propose a non-maximum removal graph contrastive learning approach (nmrGCL), 
which removes ``prominent'' dimensions (i.e., contribute most in similarity measurement) for positive pair in the pretext task. 
Comprehensive experiments on various benchmark datasets are conducted to demonstrate the effectiveness of nmrGCL, and the results show that our model outperforms the state-of-the-art methods. 
Source code will be made publicly available.
\end{abstract}

\section{Introduction}
\vspace{-3pt}
Graph Representation Learning (GRL) has become increasingly popular for the ubiquitous graph-structured data across domains, including traffic \cite{stgcn}, social network \cite{social}, and knowledge graph \cite{r-gcn}.
Graph Neural Networks (GNNs) \cite{gcn,gin}  are utilized as backbones of GRL to learn low-dimensional embeddings of nodes or graphs while maintaining structure and attribute information. 
Most GNN models are trained in the (semi-)supervised learning setting requiring abundant manually-annotated labels. 
In case of insufficient data labels, recent Contrastive Learning (CL) based on Information Maximization (InfoMax) principle \cite{infomax} has shown promising performance for self-supervised learning with success across fields including computer vision \cite{simclr,moco} and natural language processing \cite{yang-etal-2021-xmoco,gao2021simcse}. 
These CL methods maximize the Mutual Information (MI) between different augmented views of the same instance while minimizing the MI between those of the different instances. 

Inspired by the above CL models, Deep Graph InfoMax (DGI) \cite{dgi} applies the InfoMax principle to graph representation learning, which relies on maximizing the mutual information between one graph's patch-level and global-level representations. 
Following SimCLR \cite{simclr}, a series of graph contrastive learning methods  \cite{mvgrl,graphcl}  enforce the embedding of positive pair (i.e., augmented views of the same graph) to be close and the embedding of negative pair (i.e., augmented views of different graphs) to be distant in the Euclidean space. 
GCC \cite{gcc} referring to MoCo \cite{moco} contrasts graph-level embedding with momentum encoder and maintain the queue of data samples.

\begin{figure}[tb]
% \centering
\begin{minipage}[b]{0.7\textwidth} % or '[b]', if desired
    \subfigure[Complete Collapse]{\label{fig:complete}\includegraphics[width=.32\textwidth]{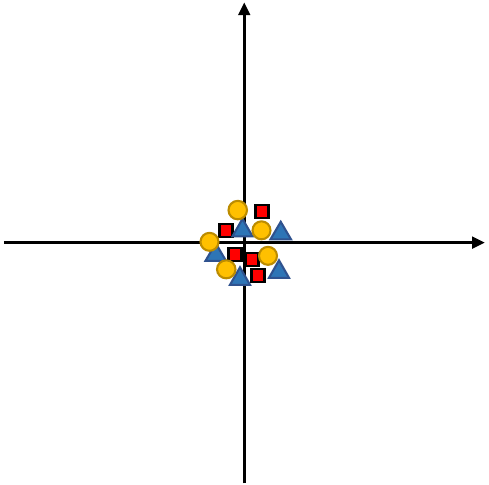}}
    \subfigure[Dimension Collapse]{\label{fig:dimensional}\includegraphics[width=.32\textwidth]{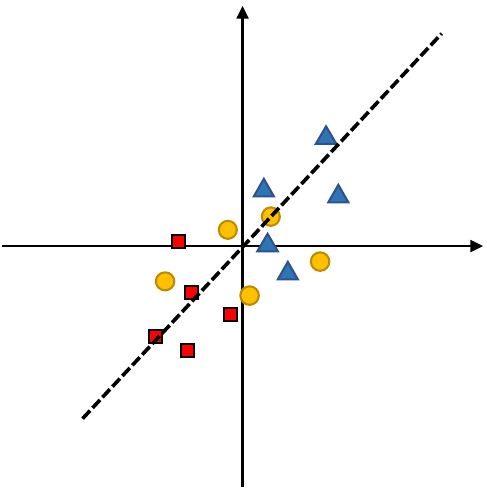}}
    \subfigure[No Collapse]{\label{fig:b}\includegraphics[width=.32\textwidth]{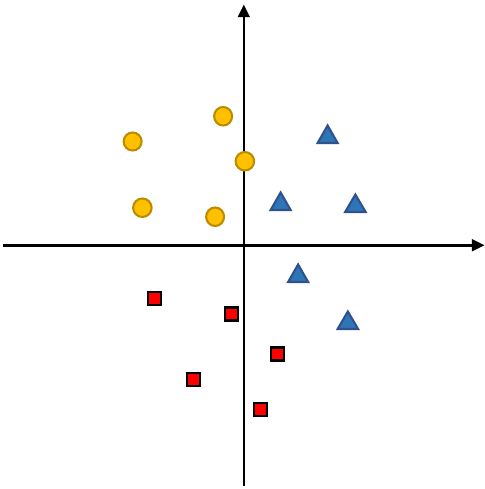}}
    \vspace{-10pt}
    \caption{Collapse Patterns in Graph Contrastive Learning.}
    \label{fig:collapse}
\end{minipage}
\begin{minipage}[b]{.28\textwidth}
    \captionsetup{width=1.1\linewidth}
    \includegraphics[width=.9\textwidth]{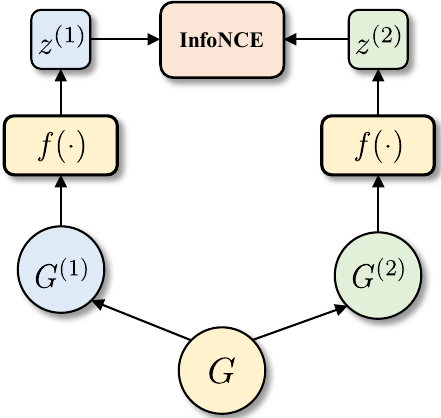}
    \caption{Pipeline of existing graph contrastive learning.}
    \label{fig:gcl} 
\end{minipage}
\vspace{-15pt}
\end{figure}

However, graph contrastive learning suffers from the dimensional collapse problem, i.e., the space spanned by embedding vectors is only a subspace of the entire space as shown in Fig.~\ref{fig:dimensional}. 
The concentration of information on part of dimensions weakens the distinguishability of embeddings in downstream classification tasks. 
We analyze two reasons for the dimensional collapse in GCL: 
(1) The smoothing effect of graph pooling layer makes initially untrained embeddings of positive pairs highly similar in pattern. 
The alignment between positive pair embeddings lead part of dimensions to collapse.
(2) The implicit regularization of graph convolution layers where the product of learnable weight matrices tend to be low-rank subsequently cause the rank deficiency of embedding space.

In this paper, we propose a novel \textbf{N}on-\textbf{M}aximum \textbf{R}emoval \textbf{G}raph \textbf{C}ontrastive \textbf{L}earning (nmrGCL) approach for self-supervised graph representation learning. 
The key idea of nmrGCL is to learn complementary embeddings of augmented graphs, inspired by non-maximum suppression (NSM) which is widely used in visual object detection \cite{zhang2018adversarial,non-max}.
Specifically, for positive pairs, the embedding of the first augmented view identifies the prominent dimensions and then removes these dimensions in the embedding of the second in pre-train process.
We conduct experiments on bioinformatics and social networks datasets to show the effectiveness. \textbf{The contributions of this paper are as follows:}

\bb $\quad$ We formally point out and theoretically analyze the so-called dimensional collapse in graph contrastive learning, which has not been discussed in graph learning literature to our best knowledge.
It limits the expressiveness of embeddings in downstream tasks e.g. node/graph classification. 
We reveal the two reasons for the dimensional collapse in GCL: i) smoothing effect of permutation-invariant graph pooling and ii) tendency to be low-rank of multi-layer graph convolution.

\bb $\quad$ We propose a novel non-maximum removal graph contrastive learning approach (nmrGCL) for self-supervised graph classification tasks and extension to node classification task. 
Our model effectively encourages the encoder to learn complementary representation in pretext tasks using non-maximum removal operation.  

\bb $\quad$ Experiments on multiple datasets show that nmrGCL outperforms state-of-the-art SSL methods on graph classification tasks and achieves competitive performance on node classification tasks.

\section{Preliminary}
\vspace{-3pt}
\textbf{Problem definition and notations.} Let $G = (V, E)$ denote a  graph, where $V =\{v_1, v_2, \cdots , v_N\}$, $E \in V \times V$ denote the node set and the edge set, respectively. 
The adjacency matrix containing the connectivity of nodes is denoted as $\mathbf{A} \in \{0,1\}^{N \times N}$, where the entry $\mathbf{A}_{ij} = 1$ if $(v_i, v_j) \in E$. 
The feature matrix is denoted as $\mathbf{X} \in \mathbb{R}^{N\times F}$, where the $i$-th column $\mathbf{x}_i \in \mathbb{R}^F$ is the $F$-dimensional feature vector of node $v_i$. 
For self-supervised graph-level representation learning, given a set of graphs $\mathcal{G} = \{G_1, G_2, \cdots\}$ without class information, our objective is to learn a GNN encoder $g_\theta(\mathbf{X}, \mathbf{A}) \in \mathbb{R}^{F^\prime}$ which encodes each graph $G$ into a $F^\prime$-dimensional vector $\mathbf{z}_G \in \mathbb{R}^{F^\prime}$. 
Self-supervised node-level representation is defined similarly.
These low-dimensional embeddings can be used in downstream tasks, such as node and graph classification. 
lv

\textbf{Graph Contrastive Learning.} Transferred from contrastive learning in the vision field, graph contrastive learning aims at learning embeddings for nodes or graphs through maximizing the consistency between augmented views of the input graph via contrastive loss. Here briefly review the pipeline of graph contrastive learning as shown in Fig.~\ref{fig:gcl}:

\textbf{i) Graph Augmentation.} 
Given a graph $G$, we define the $a$-th augmented view as $\tilde{G}^{(a)} = t_a(G)$, where $a \in \{1,2\}$, $t_a$ is selected from a group of predefined graph augmentation $\mathcal{T}$.  
Motivated by image augmentations, various graph augmentation are proposed and categorized into two types: \textit{structure-based} and \textit{feature-based}. 
Commonly used graph augmentation methods include:
1) \textbf{NodeDrop} and \textbf{NodeShuffling}: randomly discards or shuffles certain portion of nodes with their edges and features. 
2) \textbf{EdgeAdd} and \textbf{EdgeDrop}: randomly adds or drops certain portion of edges in graphs. 
3) \textbf{FeatureMasking}: randomly masks a portion of dimensions in node features with zero. 
4) \textbf{FeatureDropout}: randomly masks features of some nodes.
5) \textbf{Subgraph}: generates a subgraph with Random Walk.
The details for the used augmentation approaches are given in Appendix~\ref{app:augmentation}.

\textbf{ii) Graph Encoder.}
A shared GNN \cite{gcn,gin} encoder $f(\cdot)$ is used to extract the low-dimensional graph or node representations for each augmented graph $\tilde{G}^{(a)}$. 
Given an augmented graph $\tilde{G}^{(a)} $ with an adjacency matrix $\mathbf{A}$ and a high-dimensional node features matrix $\mathbf{X}$, where $\mathbf{x}_n = \mathbf{X}[n, :]^\top$ is the feature vector of node $v_n$, the $l$-th layer updates each node's representation by message passing:
\begin{equation}
\begin{aligned}
&\mathbf{a}_{n}^{(l)}=\texttt{AGGREGATE}^{(l)}\left(\left\{\mathbf{h}_{m}^{(l-1)}: v_{m} \in \mathcal{N}\left(v_{n}\right)\right\}\right), \\
&\mathbf{h}_{n}^{(l)}=\texttt {COMBINE}^{(l)}\left(\mathbf{h}_{n}^{(l-1)}, \mathbf{a}_{n}^{(l)}\right),
\end{aligned}
\end{equation}
where $\mathbf{h}_n^{(l)}$ is the representation of the node $v_n$ in the $l$-th layer of GNN with $\mathbf{h}_n^{(0)}=\mathbf{x}_n$, $\mathcal{N}\left(v_{n}\right)$ is the set of neighbors of node $v_n$, 
$\texttt{AGGREGATE}^{(l)}(\cdot)$ can be the sum or average operation, 
and $\texttt{COMBINE}^{(l)}(\cdot)$ can be the concatenation or average operation. 
Then the graph-level embedding of $G^{(a)}$ can be obtained through the \texttt{READOUT} function of GNN, which is similar to the pooling in CNN:
\begin{equation}
    \mathbf{r}_i = \texttt{READOUT}(\{\mathbf{h}_n^{k-1}:v_n\in V, k=1,2,\cdots,K\}),
\end{equation}
where $K$ is the number of layers of the GNN model. In this paper, a two-layer MLP is applied on top of GNN encoder to obtain $\mathbf{z}_i$. 

\textbf{iii) Embedding Contrast.} Attract positive pairs and repulse negative pairs simultaneously with InfoNCE loss by identifying the positive pair from a mini-batch.
For a mini-batch $\mathcal{B}=\{G_i^{(a)} | i\in \{1, \cdots, B\}, a \in \{1,2\}\}$, the cross-entropy loss $\mathcal{L}_i^{(a)}$ for each augmented view $G_i^{(a)}$ is:
\begin{equation}
 \mathcal{L}_{i}^{(a)} = -\log\frac{\exp{(\langle \mathbf{z}_i^{(1)}, \mathbf{z}_i^{(2)}\rangle/\tau)}}{\exp{(\langle \mathbf{z}_i^{(1)}, \mathbf{z}_i^{(2)}\rangle/\tau)} + \sum_{l\in\{1,2\}, j\in\{1,\cdots ,N\}, j\neq i}\exp{((\langle \mathbf{z}_i^{(a)}, \mathbf{z}_j^{(l)}\rangle/\tau))}}, 
 \label{eq:infonce}
\end{equation}
where $\tau$ is the temperature parameter, and the overall loss is $\mathcal{L} = \sum_{a\in\{1,2\}, i\in\{1, \cdots, N \}} \mathcal{L}_i^{(a)}$. 
\section{Understanding Dimensional Collapse in Graph Contrastive Learning}
\label{sec:dimensional_collapse}
\vspace{-3pt}
Self-supervised learning aims at learning embeddings for input instance by maximizing the similarity between embeddings of augmented views. 
These methods could fall into a collapse problem, where all input data are encoded as the same embedding. 
Contrastive methods address this problem by minimizing the distance between negative samples. 
Although graph contrastive learning which is transferred from vision field avoids complete collapse, GCL still suffers from \textbf{dimensional collapse}.
\begin{definition}[Phenomenon of Dimensional Collapse in Contrastive Learning]
Given a group of graphs $\{G_i\}$ with their embeddings $\{\mathbf{z}_i\in\mathbb{R}^{F^\prime} \}$ by certain methods, \textbf{dimensional collapse} is a phenomenon that the space spanned by the embeddings $\{\mathbf{z}_i\}$ is a proper subspace of $\mathbb{R}^{F^\prime}$.
\vspace{-5pt}
\end{definition}
We analyze that both permutation-invariant \textbf{pooling layer} and \textbf{convolution layer} as commonly adopted in existing massage-passing GNN encoders can cause dimensional collapse in GCL. 
% where the obtained embedding for graphs only span a lower-dimensional subspace. 
\subsection{Dimensional Collapse Caused by Pooling}
\begin{wrapfigure}{r}{0.45\textwidth}
\centering
\vspace{-10pt}
\includegraphics[width=0.45\textwidth]{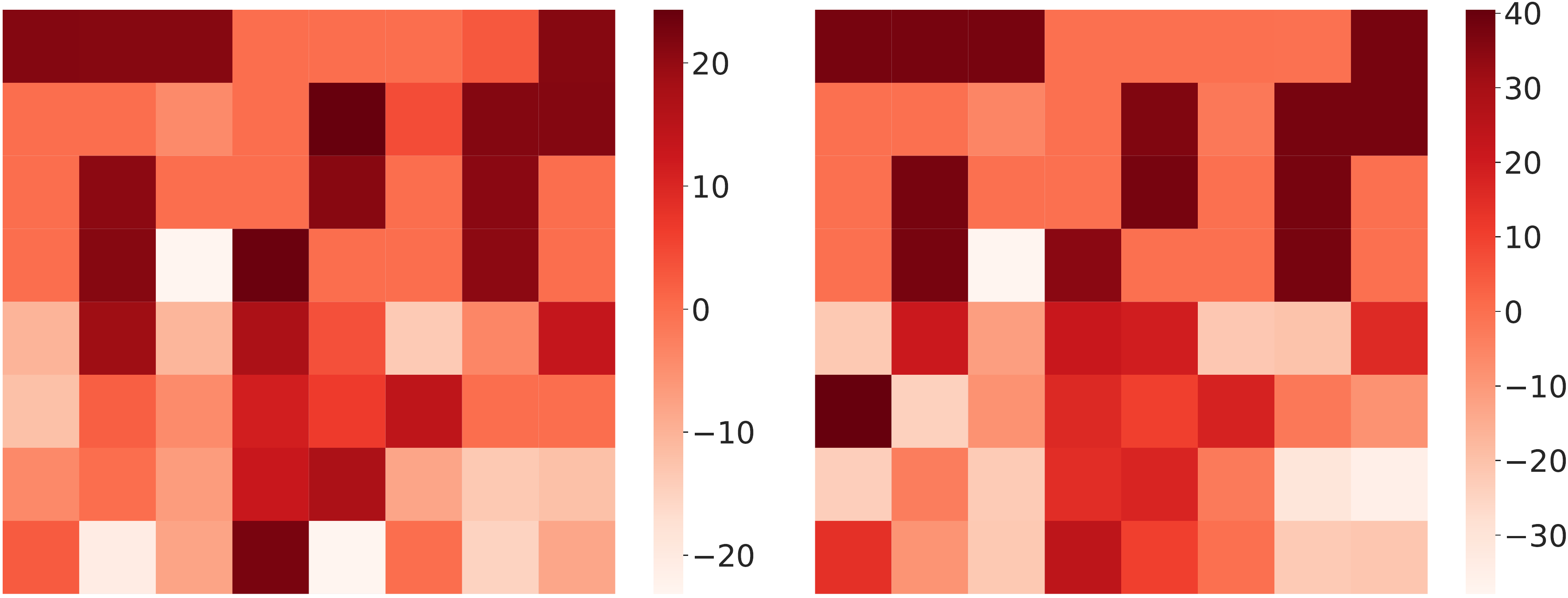}
% \vspace{-15pt}
\caption{Heat map of the 64-dim embeddings of the two augmented views from the same graph in COLLAB extracted by non-trained encoder.}
\vspace{-15pt}
\label{fig:embedding} 
\end{wrapfigure}

\vspace{-3pt}
Frequently used graph pooling layers aggregate node embeddings to obtain the graph-level embeddings with permutation invariant function including mean or summation operation for graph classification. 
We first analyze that the variance provided by augmentation will be reduced by the pooling layer, which cause dimensional collapse. 
We consider a single layer Vanilla GCN \cite{gcn} with average pooling. 
% For a graph  $G = (V, E, \mathbf{A})$, the node feature matrix is denoted as $\mathbf{X}$. 
Suppose that each graph generates two augmented views by randomly dropping $p$ portion of nodes with their features. 
Then the embedding vector for each view $G^{(a)}, a\in\{1,2\}$ is:
\begin{equation}
    \mathbf{z}_i^{(a)} = \frac{1}{|N|} \vec{e} \left(\sigma (\widetilde{\mathbf{A}}^{(a)} \mathbf{X}^{(a)} \mathbf{W}) \right),\label{eq:emb}
\end{equation}
where $\vec{e}$ is a vector of $|N|$ number of ones, $\widetilde{\mathbf{A}}^{(a)} = \mathbf{D}^{1/2}\mathbf{A}^{(a)}\mathbf{D}^{1/2}$ is the normalized adjacency matrix, $\mathbf{D}$ is the diagonal degree matrix,  $\mathbf{W}$ is the learnable parameter matrix of GCN, and $\sigma(\cdot)$ is the non-linear activation function. 
In vision field, the encoder can filter the disturbance of augmentation and extract common high-level semantic information of image. 
But in the graph field, since the graph convolution and average pooling can smooth both node and graph embeddings, initial embeddings of positive pairs extracted by arbitrary encoders are close to each other. 
We begin the analysis by considering a cycle graph and neglecting the activation function for simplicity. Also assume that the element of one-hot encoding node features follows independent identical Bernoulli distribution.
\begin{lemma}
\label{lemma:common_nodes}
Given a graph $G$ with the set of nodes $|V|=N$, two augmented views are generated by randomly discarding $p$-proportion nodes. The expected number of nodes shared by the two augmented graph is $(1-p)^2\cdot N$, i.e., $\mathbb{E}[|V^{(1)}\cap V^{(2)}|] = (1-p)^2\cdot N $
\end{lemma}
\begin{theorem} 
Given a cycle graph $G=(V,E)$ and node drop portion $p$, embeddings of two augmented views indexed by $1$ and $2$ are obtained by Eq.~\ref{eq:emb} with a random encoder. 
Then we have the lower bound for expectation of cosine similarity:
% \vspace{-10pt}
\begin{equation}
    \mathbb{E}[\texttt{sim}(\mathbf{z}^{(1)}, \mathbf{z}^{(2)})] > \frac{n+2}{(\sqrt{n}+1)^2}
\end{equation}
% \vspace{-5pt}
where $n$ is the ratio of common and unique nodes in two augmented views with expectation $\mathbb{E}[n]=(1-p)/p$ discussed in Lemma~\ref{lemma:common_nodes}. \label{theorem:similarity}
\end{theorem}
The proof is given in Appendix~\ref{app:proof_similarity}. 
It shows that, even with a random encoder, the difference between embeddings of positive pairs are small. 
Similar results can be obtained when using other augmentation methods. 
% The growth in any of $n$, $|V|$, $|E|$ will intensify this situation as shown in the proof. 
We discuss Theorem~\ref{theorem:similarity} with a toy example. 
Consider a cycle graph $G$ with 5 nodes connected sequentially.  
The feature vector of each node is $\{\mathbf{x}_1, \mathbf{x}_2, \mathbf{x}_3, \mathbf{x}_4, \mathbf{x}_5\}$. 
Two augmented views are generated by dropping the 4-th and 5-th node respectively. 
The graph convolution layer aggregates the feature of neighbors and they becomes: 
% $\{(\mathbf{x}_5+\mathbf{x}_1+\mathbf{x}_2)\cdot \mathbf{W} /3, (\mathbf{x}_1+\mathbf{x}_2+\mathbf{x}_3)\cdot \mathbf{W}/3, (\mathbf{x}_2+\mathbf{x}_3)\cdot \mathbf{W}/2, (\mathbf{x}_5+\mathbf{x}_1)\cdot \mathbf{W}/2\}$ and 
% $\{(\mathbf{x}_1+\mathbf{x}_2)\cdot \mathbf{W}/2, (\mathbf{x}_1+\mathbf{x}_2+\mathbf{x}_3)\cdot \mathbf{W}/3, (\mathbf{x}_2+\mathbf{x}_3+\mathbf{x}_4)\cdot \mathbf{W}/3, (\mathbf{x}_3+\mathbf{x}_4)\cdot \mathbf{W}/2$.
$\{(\mathbf{x}_5/\sqrt{2}+\mathbf{x}_1/2+\mathbf{x}_2/2)\cdot \mathbf{W}, (\mathbf{x}_1/2+\mathbf{x}_2/2+\mathbf{x}_3/\sqrt{2})\cdot \mathbf{W}, (\mathbf{x}_2/\sqrt{2}+\mathbf{x}_3)\cdot \mathbf{W}, (\mathbf{x}_5+\mathbf{x}_1/\sqrt{2})\cdot \mathbf{W}\}$ and 
$\{(\mathbf{x}_1+\mathbf{x}_2/\sqrt{2})\cdot \mathbf{W}, (\mathbf{x}_1/\sqrt{2}+\mathbf{x}_2/2+\mathbf{x}_3/2)\cdot \mathbf{W}, (\mathbf{x}_2/2+\mathbf{x}_3/2+\mathbf{x}_4/\sqrt{2})\cdot \mathbf{W}, (\mathbf{x}_3/\sqrt{2}+\mathbf{x}_4)\cdot \mathbf{W}\}$.
The graph-level embeddings of augmented views obtained with average pooling layer are:
% \begin{small}
\begin{equation*}
\mathbf{z}^{(1)} =\frac{2+\sqrt{2}}{8}(\mathbf{x}_1 + \mathbf{x}_2 +\mathbf{x}_3 + \mathbf{x}_5)\cdot \mathbf{W},\quad
\mathbf{z}^{(2)} =\frac{2+\sqrt{2}}{8}(\mathbf{x}_1 + \mathbf{x}_2 +\mathbf{x}_3 + \mathbf{x}_4)\cdot \mathbf{W}.\\
\end{equation*}
% \end{small}
% \vspace{-20pt}
% \vspace{-20pt}
% Although the two augmented graph are generated by dropping 20\% nodes, 
The difference between embeddings is mainly attributed to the different retained nodes, which makes the two embeddings highly similar.
This situation also occurs when using other augmentation methods and considering activation function.
We empirically demonstrate this property through a specific experiment. 
We randomly select a graph from COLLAB dataset and generate two augmented views by dropping 30\% nodes and masking 30\% dimensions of node features, respectively. 
Then generate two 64-dimensional embeddings with a randomly initialized 3-layer GCN with ReLU activation. Shown as the heat map in Fig.~\ref{fig:embedding}, the patterns of the two embeddings strongly aligns.

Additionally, we observe that parts of elements are much larger than others in both embeddings and these larger elements share the same positions. 
In this paper, we introduce a threshold $\delta$ according which, we define the position of elements larger than $\delta$ as \textbf{``prominent'' dimensions}. 
\begin{definition}
Given a graph $G$ with embedding $\mathbf{z}\in\mathbb{R}^F$ and a threshold $\delta$, the \textbf{``prominant'' dimensions} are defined as: $D = \{d \mid \mathbf{z}[d] > \delta\}$, where $\mathbf{z}[d]$ is the $d$-th element of $\mathbf{z}$.
\end{definition}
% We further show that the gradient of InfoNCE with respect to positive pairs is product of a multiplier and the embedding. 
We then analyze how ``prominent'' dimensions cause the dimensional collapse from the perspective of the gradient of InfoNCE with derivation in Appendix~\ref{app:proof_g_z}.
The gradient of InfoNCE w.r.t the embeddings of positive pairs are: 
\begin{equation}
\left\{\begin{array}{l}
% -\nabla_{\mathbf{z}_{i}^{(1)}} L_{i}^{(1)}=\frac{q_{B, i}^{(1)}}{\tau}\left[\mathbf{z}_{i}^{(2)}-\sum_{l \in\{1,2\}, j \in \llbracket 1, N \rrbracket, j \neq i} \frac{\exp \left\langle\mathbf{z}_{i}^{(1)}, \mathbf{z}_{j}^{(l)}\right\rangle / \tau}{\sum_{q \in\{1,2\}, j \in \llbracket 1, N \rrbracket, j \neq i} \exp \left(\left\langle\mathbf{z}_{i}^{(1)}, \mathbf{z}_{j}^{(q)}\right\rangle / \tau\right)} \cdot \mathbf{z}_{j}^{(l)}\right] \\
\frac{\partial \mathcal{L}_{i}^{(1)}}{\partial \mathbf{z}_i^{(1)}}=-\frac{1}{\tau}
\left(\left(1- \frac{\exp(\langle \mathbf{z}_i^{(1)} , \mathbf{z}_i^{(2)} \rangle)}{p_i}\right) \cdot \mathbf{z}^{(2)}_i- \sum_{l\in\{1,2\}, j\in\{1,2,\cdots,N\},j\neq i}\left(\frac{\exp(\langle \mathbf{z}_i^{(1)}, \mathbf{z}_j^{(l)} \rangle)}{p_i}\cdot \mathbf{z}^{(l)}_j\right)\right)\\
% -\nabla_{\mathbf{z}_{i}^{(2)}} L_{i}^{(1)}=\frac{q_{B, i}^{(1)}}{\tau} \cdot \mathbf{z}_{i}^{(1)} \\
\frac{\partial \mathcal{L}_{i}^{(1)}}{\partial \mathbf{z}^{(2)}_i} = - \frac{1}{ \tau}\left(1- \frac{\exp(\langle \mathbf{z}_i^{(1)} , \mathbf{z}_i^{(2)} \rangle)}{p_i} \right)\cdot \mathbf{z}^{(1)}
% -\nabla_{\mathbf{z}_{j}^{(l)}} L_{i}^{(1)}=-\frac{q_{B, i}^{(1)}}{\tau} \frac{\exp \left\langle\mathbf{z}_{i}^{(1)}, \mathbf{z}_{j}^{(l)}\right\rangle / \tau}{\sum_{q \in\{1,2\}, j \in \llbracket 1, N \rrbracket, j \neq i} \exp \left(\left\langle\mathbf{z}_{i}^{(1)}, \mathbf{z}_{j}^{(q)}\right\rangle / \tau\right)} \cdot \mathbf{z}_{i}^{(1)}
\end{array}\right. \label{eq:g_z}
\end{equation}
where 
$
p_i =
{\exp{(\langle \mathbf{z}_i^{(1)},  \mathbf{z}_i^{(2)}\rangle/\tau)} +  \sum_{l\in\{1,2\}, j\in\{1,\cdots,N\},j\neq i} \exp (\langle \mathbf{z}_i^{(1)}, \mathbf{z}_j^{(l)}\rangle/\tau)}
$.
Note that the gradient of $\mathcal{L}_i^{(1)}$ w.r.t $\mathbf{z}_i^{(1)}$ and $\mathbf{z}_i^{(2)}$  grow linearly with the value of $\mathbf{z}_i^{(2)}$ and $\mathbf{z}_i^{(1)}$ in each dimension, which still holds for $\mathcal{L}_{i}^{(2)}$ due to the symmetry. 
Also because two embeddings share a number of common prominent dimensions, minimization of InfoNCE loss mainly lies in increasing prominent dimensions which will be more prominent cumulatively. 
Intuitively, contrastive learning aiming at maximizing the similarity between positive pairs leads to a shortcut that only a few dimensions to be relatively much larger. 
These prominent dimensions control the representation of graphs and suppress the expressiveness of other dimensions.
Consequently, the downstream classification performance mainly depends on prominent dimensions and neglects the leverage of other dimensions.

\subsection{Dimensional Collapse Caused by Graph Convolution Layer}
\vspace{-3pt}
% We then analyze another cause of dimensional collapse, which is the implicit regularization of the graph convolution layer. 
% As a form of implicit regularization, alignment phenomena \cite{ji2018gradient,ji2020directional} is that the direction of gradient of the loss converges to the direction of gradient flow path.
% Subsequently, \cite{dimensional_collapse} shows that the over-parameterized MLP in SimCLR \cite{simclr} tend to generate low-rank embeddings. 
% Since massage passing graph neural network can be considered as the MLP multiplied by an adjacency matrix, we follow the \cite{dimensional_collapse} to analyze how implicit regularization of graph neural networks cause dimensional collapse of graph contrastive learning.
% 
We then analyze the other cause of dimensional collapse in terms of an implicit regularization of the graph convolution layer, where the learnable weight matrices have tendency to be low-rank. 
The implicit regularization of neural network has been studied in \cite{implicit_regularization,gunasekar2017implicit,ji2020directional}. We consider a $K$-layers Vanilla GCN as the encoder for graph contrastive learning. 
For simplicity, we neglect the pooling layer and  the activation function. 
% and concatenate embeddings of all nodes as the graph-level embeddings.
Node embeddings are concatenated as the graph-level embeddings.
The learnable weight matrix of graph convolution layers are $\mathbf{W}_1, \mathbf{W}_2, \cdots, \mathbf{W}_K$. 
Additional, we choose feature-based augmentation methods such as Feature Masking or Feature Dropout. 
The feature matrix of two augmented views are $\mathbf{X}^{(1)}$ and $\mathbf{X}^{(2)}$ and the normalized adjacency matrix $\widetilde{\mathbf{A}}$ of the augmented views remains unchanged. 
Thus, the embedding vectors for augmented views are:
\begin{equation}
\mathbf{z}^{(a)} = (\cdots(\widetilde{\mathbf{A}}(\widetilde{\mathbf{A}}\mathbf{X}^{(a)}\mathbf{W}_1)\mathbf{W}_2)\cdots \mathbf{W}_K)=\widetilde{\mathbf{A}}^K\mathbf{X}^{(a)}\mathbf{W}_1\cdots \mathbf{W}_K,
\end{equation}
where $K$ is the depth of the GNN encoder.
Note the weight matrices can be written as product form, $\mathbf{W}:=\mathbf{W}_1\mathbf{W}_2\cdots \mathbf{W}_K$.
We study the characteristic of gradient chain (gradient descent with infinitesimally small learning rate) of the matrices product. We denote the gradient flow of $\mathbf{W}$ as:
\begin{equation}
    \dot{\mathbf{W(t)}}:= \frac{d\mathbf{W(t)}}{dt} = -\frac{\partial \mathcal{L} }{\partial \mathbf{W(t)}}, \quad t\geq0, 
\end{equation}
where $t$ is the continuous time index since the infinitesimally small learning rate. 
We first perform the singular value decomposition (SVD) on matrices product:
\begin{equation}
\mathbf{W}(t) = \mathbf{U}(t) \mathbf{\Sigma(t)} \mathbf{V}^\top(t), \label{eq:svd}
\end{equation}
such that $\mathbf{U}$ and $\mathbf{V}$ are orthonormal matrices, $\mathbf{\Sigma}$ are diagonal matrices whose $m$-th diagonal entry equals the $m$-th singular value $\sigma_m$. 
The gradient flow of singular values is simply derived in Appendix~\ref{app:proof_g_sigma_lemma} given that $\mathbf{U}$ and $\mathbf{V}$ have orthonormal columns:
\begin{lemma}\label{lemma:g_sigma}
The derivative of singular values of the matrix product $\mathbf{W}$ w.r.t time are:
\begin{equation}
    \dot{\sigma_m}(t) = \mathbf{u}_m^\top(t) \dot{\mathbf{W}}(t) \mathbf{v} (t),\label{eq:g_sigma}
\end{equation}
where $\sigma_m$ is the $m$-th singular value of $\mathbf{W}$, $\mathbf{u}_m$ and $\mathbf{v}_m$ is the $m$-th column of $\mathbf{U}$ and $\mathbf{V}$, respectively.
\end{lemma}
We adopt the expression for the gradient flow of the matrices product from Theorem~1 in \cite{arora2018optimization} as:
\begin{equation}
    \dot{\mathbf{W}(t)} = \sum_{j=1}^N \left[\mathbf{W}^\top(t)\mathbf{W}(t)\right]^\frac{K-j}{K}\cdot\frac{\partial \mathcal{L}(\mathbf{W}(t))}{\partial\mathbf{W}} \cdot \left[\mathbf{W}(t)\mathbf{W}^\top(t)\right]^\frac{j-1}{K}, \label{eq:g_w}
\end{equation}
where $[\cdot]^\alpha, \alpha \in \mathbb{R}^+$ denotes power operator over positive semi-definite matrices. 
% With Lemma~\ref{lemma:g_w} and Lemma~\ref{lemma:g_sigma}, 
Substituting Eq.~\ref{eq:g_w} into Eq.~\ref{eq:g_sigma}
we can analyze the gradient flow of singular values under InfoNCE loss. 
\begin{theorem}\label{theorem:g_sigma}
The singular values of the product matrix $W(t)$ evolve by:
\begin{equation}
\dot{\sigma_m}(t) = -K\cdot (\sigma_m^2 (t))^{\frac{K-1}{K}} \cdot \mathbf{u}_m^\top(t) \cdot \mathbf{C_z} \cdot\mathbf{v}_m^{(t)},
\end{equation}
where $\mathbf{C_\mathbf{z}} = \sum_i (\mathbf{g}_{\mathbf{z}_i^{(1)}} \widetilde{\mathbf{A}}^K\mathbf{X}_i^{(1)} + \mathbf{g}_{\mathbf{z}_i^{(2)}}\widetilde{\mathbf{A}}^K\mathbf{X}_i^{(2)}) $, and $g_{\mathbf{z}_i^{(k)}}$ is the gradient over $\mathbf{z}_i^{(k)}$ in Eq.~\ref{eq:g_z}. 
\end{theorem}
The proof is in Appendix~\ref{app:proof_g_sigma}. 
Theorem~\ref{theorem:g_sigma} states that the gradients flow of the $m$-th singular value $\dot{\sigma_m}$ for matrices product of graph convolution layer are proportional to the value of that singular value $\sigma_m$.
The larger singular values grow considerably faster than smaller ones, thus some of singular value of weight matrices turn out to be relatively small and close to zero.
Moreover, singular value decomposition factorize a matrix into a combination of a group of orthonormal bases. 
The singular values represent scale of corresponding bases while the number of non-zero singular values equals to the rank of the matrix.
% The rank of a matrix equals to the number of non-zero singular values. 
Thus, the weight matrices product tend to be approximately \textbf{low-rank}.
The embedding space is applying a linear transformation (weight matrices product) on the feature space.  
The rank deficiency of matrices product finally leads the embedding space to be low-rank, i.e., dimensional collapse of embedding space.

\begin{figure*}[tb!] 
\centering
\includegraphics[width=0.9\textwidth]{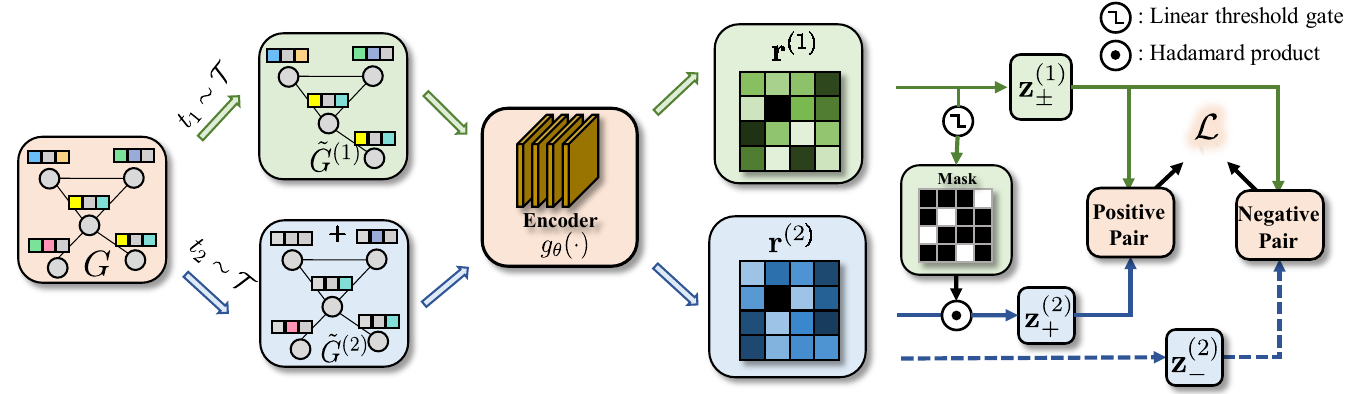}
\vspace{-10pt}
\caption{Overview of nmrGCL which here involves two augmented views of each input graph. 
A shared GNN encoder $g_\theta(\cdot)$ encodes each augmented view as low-dimensional embeddings, where the darker colors represents the larger values. 
% The second embedding in positive pair is performed with removal operation where the ``prominent'' dimensions of the first embedding is masked with zeros.
For positive pairs, the first embedding identifies prominent dimensions whose values are larger than a predefined threshold, and masks these dimensions of the second embedding with zeros.
% which complement the first.
% which is reffe complementary information.
% The second embedding as the complementary embedding is leveraged to extract hard-to-learn information. 
% Negative pairs are simply the embeddings of augmented views of different graphs.
The removal operation is not performed on negative pairs. 
% Finally, the parameters of the shared encoder $\theta$ are optimized by minimizing the objective function with gradient descent.
} 
\vspace{-13pt}
\label{fig:overview} 
\end{figure*}
% \textbf{Motivations.} 

\section{Proposed Approach}
\vspace{-3pt}
As discussed in Sec.~\ref{sec:dimensional_collapse}, current graph contrastive learning paradigm suffers the dimensional collapse, where parts of embedding dimension lose efficacy in representing the input graph.
As illustrated in Fig.~\ref{fig:dimensional}, when the embeddings are compressed into a lower-dimensional subspace, samples with different labels are hard to separate along the axis.
Henceforth, the obtained embeddings may suffer underfitting in downstream tasks e.g. graph classification. 
More concretely speaking, we note that \cite{Wasserstein_discriminant_analysis,9008561,hou2016squared} quantify the inter-class separability with Wasserstein distance between different classes. 
% To enhance the inter-class separability 
In fact, by increasing the distance between samples on collapsed dimensions, Wasserstein distance between classes can be increased, indicating stronger inter-class separability.
Based on these evidences and analysis, 
% to narrow the gap among dimensions, \
we propose the non-maximum removal graph contrastive learning (nmrGCL) which directly set prominant dimensions to zero for one of positive pair. 

As shown in Fig.~\ref{fig:overview}, given the input graph $G$, we first generate two augmented graph views $\tilde{G}_1$ and $\tilde{G}_2$. 
Then each augmented view is encoded into a low-dimensional embedding with one shared encoder. 
After that, for positive pairs (i.e., two augmented views of the same graph), the embedding of the first augmented view conducts the removal operation on the embedding of the second one,  to prevent the shortcut that only focuses on part of dimensions to maximize the similarity. 
Since the negative pairs (i.e., augmented views of the different graph) do not share the same prominent dimensions, the removal operation is not applied on the negative pairs. 
Finally, the parameters of the encoder are learned with contrastive objectives.

\textbf{Erasing Operation.}
The key innovation of our proposed nmrGCL approach is the removal operation which enforces the encoder to learn complementary embeddings.
nmrGCL aims at mining information of inconspicuous dimensions rather than being restricted to scarce common prominent dimensions through an adversarial manner. 
After the GNN encodes two augmented views $\{\tilde{G}^{(1)}, \tilde{G}^{(2)}\}$ into embeddings $\{\mathbf{z}^{(1)}, \mathbf{z}^{(2)}\}$, 
the embedding of the second augmented view $\mathbf{z}^{(2)}$ is erased with the guidance of
% the embedding of the first augmented view 
$\mathbf{z}^{(1)}$.  
Formally, the augmented views of the input graph are transformed by the GNN encoder $g_\theta (\cdot)$ into a pair of embeddings $r^{(a)}\in \mathbb{R}^{K\times P}$, where $a\in\{1,2\}$, $K$ is the number of layers of GNN, and $P$ is the number of output dimensions in each layer. 
A projection head $h(\cdot)$ composed of a 2-layer MLP and ReLU non-linearity is applied on all embeddings for optimization objective, 
$
\mathbf{z}^{(1)} = h(\mathbf{r}^{(1)}), \mathbf{z}^{(2)}  = h(\mathbf{r}^{(2)}).
$
$\mathbf{z}^{(1)}$ is normalized to $[0, 1]$ with min-max scaling which is denoted as $\overline{\mathbf{z}}^{(1)}$.
Then a binary mask $\mathbf{M}\in\{0,1\}^{F^\prime}$ is created to conduct the removal operation on the embedding of the second augmented view $\mathbf{z}^{(2)}$ by:
% where the same dimensions as the highlighted dimensions of $r^{(1)}$ are set to 1 and others are set to 0. 
\begin{equation}
\mathbf{M}[i]= \begin{cases}0, & \text { if }  \overline{\mathbf{z}}^{(1)}[i]>\delta \\ 1, & \text { otherwise }\end{cases}
\label{eq:mask}
\end{equation}
where the threshold $\delta$ is a hyper-parameter. 
% The erased embedding of the second augmented view $\hat{r}^{(2)} =  $ is obtained by element-wisely multiplying $r^{(2)}$ with the mask matrix. 
The erased embedding $\hat{\mathbf{z}}^{(2)}$ as the complementary embedding can be obtained with
$
\hat{\mathbf{z}}^{(2)} = \mathbf{z}^{(2)} \odot \mathbf{M},
$
where $\odot$ is the Hadamard product.

% \end{tiny}
% \sum_{j^\prime=1, j^\prime\neq j}^m \exp(sim(z^{(1)}_{j,-}, z^{(2)}_{j^\prime,-}) )
% \subsection{Model Training}
\textbf{Model Training}
We  train the nmrGCL end-to-end by maximizing the agreement between positive pairs $\{\mathbf{z}_+^{(1)}, \mathbf{z}_+^{(2)}\}$ via InfoNCE loss. 
% We use the Info Noise-Contrastive Estimation (InfoNCE) \cite{infonce} which is a lower bound of the mutual information as the training objective. 
Specifically, given a mini-batch of $B$ graphs $\{G_j\}_{j=1}^{B}$, let 
$\mathbf{z}_{j,\pm}^{(1)} = h(g_\theta(t_1(G_j)))$, 
$\mathbf{z}_{j,+}^{(2)} = h(g_\theta(t_2(G_j)) )  \odot \mathbf{M}$ ,
and $\mathbf{z}_{j,-}^{(2)} = h(g_\theta(t_2(G_j)))$, where $t_a(\cdot)$ is the augmentation function, $g_\theta(\cdot)$ is the GNN encoder, $h(\cdot)$ is the projection header,
% implemented by a 2-layer MLP with ReLU activation layer, 
$\mathbf{M}$ is the mask in Eq.~\ref{eq:mask}, and $\{+, -\}$ stands for positive or negative pairs. The loss for the mini-batch with size $B$ is given as follows, where $\texttt{sim}(\cdot, \cdot)$ denotes a similarity measure e.g. cosine used in the paper:
\begin{small}
\begin{equation}
\label{eq:loss}
    \mathcal{L} = -\frac{1}{2B}\sum_{j=1}^B \sum_{a\in\{1,2\}} \log \frac{\exp(\texttt{sim}(\mathbf{z}^{(1)}_{j,+}, \mathbf{z}^{(2)}_{j,+})/\tau )}{\exp(\texttt{sim}(\mathbf{z}^{(1)}_{j,+}, \mathbf{z}^{(2)}_{j,+}) /\tau)+ \sum_{j^\prime=1, j^\prime\neq j, l\in\{1,2\}}^B\exp(\texttt{sim}(\mathbf{z}^{(a)}_{j,-}, \mathbf{z}^{(l)}_{j^\prime,-})/\tau )},
\end{equation}
\end{small}
The training process is described in Algorithm \ref{alg:nmrGCL} of Appendix~\ref{app:training_algoirthm}.

% \section{Theoretical Analysis}
% In order to portray what kind of embedding  is the optimal embedding in contrastive learning, which is most meaningful for downstream tasks, we have the following definition:
% \begin{definition}[Optimal representation in contrastive learning]
% Given a encoder $f(\cdot)$ which extract an embedding $r$ from the input data $x$, 
% \end{definition}
\section{Related Work}
\vspace{-3pt}

\textbf{Graph contrastive learning.}
As one of the main approaches of self-supervised representation learning, contrastive learning based on mutual information maximization(MI) principle has raised a surge of attraction in computer vision \cite{simclr,moco} and natural language processing \cite{yang2019reducing,li2021kfcnet}. 
% Contrastive learning is based on the mutual information (MI) maximization principle by encouraging the agreement between two augmented views of the same instance and the disagreement between augmented views of different instances. 
Inspired by visual contrastive learning, a series of graph contrastive learning methods are devised. Deep Graph InfoMax (DGI) \cite{dgi} first applies the InfoMax principle to graph representation learning. 
DGI relies on maximizing the mutual information between patch-level and global-level representation of one graph. 
GMI \cite{GMI} jointly maximizes feature MI and edge MI individually, without augmentation. 
MVGRL \cite{mvgrl} generates two augmented graph view via graph diffusion and subgraph sampling. 
Based on SimCLR \cite{simclr}, GraphCL \cite{graphcl} enforces the embedding of positive pair (i.e., augmented views of the same graph) to be close and the embedding of negative pair (i.e., augmented views of different graphs) to be distant in the  Euclidean space. 
% GCC \cite{gcc} built on MoCo \cite{moco} contrasts graph-level embedding with momentum encoder and maintain the queue of data samples. 
CuCo \cite{cuco} further utilizes the curriculum learning to select the negative samples. 
AD-GCL \cite{adgcl} optimizes adversarial graph augmentation to prevent learning redundant information. 
JOAO \cite{joao} adaptively selects the augmentation for specific dataset.

\textbf{Collapse in Self-Supervised Learning.} Self-supervised contrastive learning methods may suffer from collapse problem as shown in Fig.~\ref{fig:complete}, i.e., obtained embeddings degenerate to an constant vector. 
MoCo \cite{moco} and SimCLR \cite{simclr} address the collapse problem by repulsing negative pairs in optimization objective. 
BYOL \cite{byol} propose \textit{momentum encoder}, \textit{predictor} and \textit{stop gradient operator} to avoid the collapsed solution. 
SimSiam \cite{simsiam} simplifies the BYOL by removing the momentum encoder and shows that the remaining \textbf{stop-gradient} mechanism is the key component to prevent collapse in self-supervised learning.  
\cite{simsiam_analysis} theoretically analyze how SimSiam avoid collapse solution without negative samples. 
\cite{feature_decorrelation,dimensional_collapse} point out the dimensional collapse and further show that strong augmentation and the weight matrices alignment cause the dimensional collapse in vision field. 
% \section{Preliminaries}

\section{Experiment}
\vspace{-3pt}
\label{sec:exp}
\subsection{Protocol and Setups}
\vspace{-3pt}
\textbf{Task and datasets.} We compare state-of-the-art methods in the settings of unsupervised graph classification tasks and further extend to node classification, where we can only access the adjacency matrix and node features of graphs without label information. We conduct experiments on nine benchmark datasets \cite{tudataset}, including four bioinformatics datasets (PTC-MR, PROTEINS, NCI1, DD) and five social network datasets (COLLAB, IMDB-BINARY, IMDB-MULTI, REDDIT-BINARY, and REDDIT-M5K) for graph classification tasks and four datasets (Pubmed, Coauthor-CS, Amazon-Photo, Amazon-Computer) for node classification tasks. Datasets details are in Appendix~\ref{app:datasets}.

% \subsubsection{Baselines}

\textbf{Baselines.} We compare with three groups of baselines for graph classification. The first group is supervised GNNs including \textbf{GCN} \cite{gcn} and \textbf{GIN} \cite{gin}. The second group includes the graph kernel methods: Weisfeiler-Lehman Sub-tree kernel (\textbf{WL}) \cite{wl} and Deep Grph Kernels (\textbf{DGK}) \cite{dgk}. The last group includes unsupervised graph representation learning methods: \textbf{Sub2Vec} \cite{sub2vec}, \textbf{Graph2Vec} \cite{graph2vec}, \textbf{InfoGraph} \cite{infograph}, \textbf{GraphCL} \cite{graphcl}, \textbf{CuCo} \cite{cuco}, \textbf{AD-GCL} \cite{adgcl}, where the last four methods are state-of-the-art contrast-based graph representation learning methods. For node classification, we compare nmrGCL with supervised methods including \textbf{GCN} \cite{gcn} and \textbf{GIN} \cite{gin} and contrast-based self-supervised methods: \textbf{DGI} \cite{dgi}, \textbf{MVGRL} \cite{mvgrl}, \textbf{GCA} \cite{gca} and \textbf{CCA-SSG} \cite{ccassg}.

\textbf{Implementation Details.} 
For graph classification, we use the graph isomorphism network (GIN) \cite{gin} as the encoder for its expressiveness in distinguishing the structure to obtain the graph-level representation. 
Specifically, we adopt a three-layer GIN with 32 hidden units in each layer and a sum pooling readout function.
Then the embeddings generated by the encoder are fed into the downstream SVM classifier.
The threshold $\delta$ in the removal operation is set to $0.7$. 
We utilize the 10-fold CV to train the SVM and record mean accuracy with standard variation of five-time trials. 
Other hyper-parameters remain consistency with the GraphCL \cite{graphcl}.
For node classification tasks, we follow settings of DGI \cite{dgi} which uses the GCN as the encoder and logistic regression downstream classifier. 
More details of the experimental setup can be found in Appendix~\ref{app:exp}.
\subsection{Experimental Results} 
\label{sec:exp_result}
\vspace{-3pt}
\textbf{Main results.}
The results of self-supervised graph classification are reported in Tab.~\ref{tab:comp_result}. 
We can see contrast-based methods generally exceed both the graph kernel methods and traditional unsupervised methods, indicating the advantages of contrastive learning.
nmrGCL outperforms other unsupervised representation learning baselines with significant improvement across eight of nine datasets, especially on sparse-graph, demonstrating the superiority of our approach. 
For example, nmrGCL achieves $80.61\%$  accuracy on dense-graph dataset DD, surpassing GraphCL by $1.99\%$ accuracy and CuCo by $1.41\%$ accuracy individually. 
Meanwhile, the nmrGCL achieves $73.83\%$ accuracy on sparse-graph dataset IMDB-B, exceeding the GraphCL by $2.69\%$ and AD-GCL by $2.34\%$ accuracy, respectively. 
The results are attributed to the key component in our approach: the non-maximum removal operation, which allows information to be represented in all dimensions of embeddings rather than concentrating on a small number of the prominent dimensions. 
The similarity between two embeddings is no longer dominated only by prominent dimensions in both embeddings as in previous works. 
Finally,  with a more uniform distribution of information, all dimensions of the embeddings contribute when performing downstream classification tasks, bringing about a notable improvement. 
However, results of contrast-based methods show a large variance due to the randomness, which indicates that improving the stability of graph contrastive learning is a valuable direction.
Tab.~\ref{tab:node_classification} reports the results of node classification.
Since the GNN encoder in the graph classification task has one pooling layer which is not in the encoder for the node classification task,
% We design our approach from the gradient of InfoNCE loss w.r.t the graph-level embeddings which is the output of the readout layer.
only the weight matrices of graph convolution layer cause the dimensional collapse in node classification tasks. 
Thus our approach is more suitable for graph-level classification.
Results show that our method achieves competitive performance compared to the SOTA self-supervised node classification methods. 

% \newpage
\begin{figure}[tb!] 
\centering
\includegraphics[width=1\textwidth]{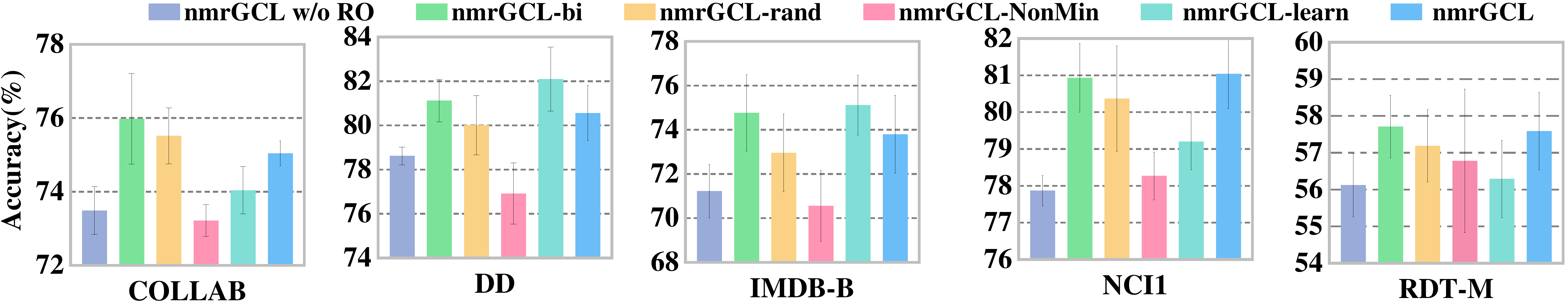}
\vspace{-15pt}
\caption{Graph classification comparison of five variants on five datasets.} 
\vspace{-15pt}
\label{fig:ablation} 
\end{figure}

\begin{table}[tb!]

  \centering
\caption{Unsupervised graph classification  performance comparison (\% with std).}
% \vspace{-10pt}
%   \scalebox{0.}{
 \resizebox{\textwidth}{22mm}{
 \renewcommand\tabcolsep{1pt}
% Table generated by Excel2LaTeX from sheet 'Sheet2'
\begin{tabular}{c|ccccccccc}
\toprule
\multicolumn{1}{c}{Datasets} & PTC-MR & PROTEINS & NCI1 & DD & COLLAB & IMDB-B & IMDB-M & RDT-B & RDT-M5K \\
\midrule
\midrule
\multicolumn{10}{c}{Supervised Methods} \\
\midrule
GCN & $64.2\pm4.3$ & $76.2\pm2.8$ & $80.2\pm2.0$ & $76.2\pm1.4$  & $79.0\pm1.8$ & $74.0\pm3.4$ & $51.9\pm3.8$ & $<50.0$  & $<20.0$ \\
GIN & $64.6\pm7.0$ & $76.6\pm3.2$ & $82.7\pm1.7$ & $78.9\pm1.3$  & $80.2\pm1.9$ & $75.1\pm5.1$ & $52.3\pm2.8$ & $92.4\pm2.5$ & $57.0\pm1.7$ \\
\midrule
\multicolumn{10}{c}{Graph Kernel Methods} \\
\midrule
WL & $57.97\pm0.49$ & $72.92\pm0.56$ & $80.01\pm0.50$ & $79.78\pm0.36$ & $69.30\pm3.42$ & $72.30\pm3.44$ & $46.95\pm0.46$ & $68.82\pm0.41$ & $46.06\pm0.21$ \\
DGK & $60.08\pm2.55$ & $73.30\pm0.82$ & $80.31\pm0.46$ & $74.85\pm0.74$ & $64.66\pm0.50$ & $66.96\pm0.56$ & $44.55\pm0.52$ & $78.04\pm0.39$ & $41.27\pm0.18$ \\
% MLG & $63.26\pm1.48$ & $41.23\pm0.04$ & -  & -  & -  & $66.55\pm0.25$ & $41.17\pm0.03$ & -  & - \\
\midrule
\multicolumn{10}{c}{Unsupervised Methods} \\
\midrule
Sub2Vec & $59.99\pm6.38$ & $53.03\pm5.55$ & $52.84\pm1.47$ & $54.33\pm2.44$ & $55.26\pm1.54$ & $55.32\pm1.52$ & $36.57\pm0.82$ & $71.48\pm0.42$ & $36,68\pm0.42$ \\
Graph2Vec & $60.17\pm6.86$ & $73.30\pm2.05$ & $73.22\pm1.81$ & $79.32\pm2.32$ & $71.10\pm0.54$ & $71.12\pm0.47$ & $50.43\pm0.88$ & $75.78\pm1.03$ & $46.86\pm0.26$ \\
InfoGraph & $61.65\pm1.43$ & $74.44\pm0.31$ & $76.20\pm1.06$ & $72.85\pm0.31$ & $70.65\pm1.13$ & $73.02\pm0.93$ & $49.65\pm9.52$ & $82.50\pm1.42$ & $53.46\pm1.03$ \\
GraphCL & $63.62\pm1.83$ & $74.39\pm0.45$ & $77.87\pm0.41$ & $78.62\pm0.40$ & $71.36\pm1.15$ & $71.14\pm0.44$ & $50.69\pm0.43$ & $89.53\pm0.84$ & $55.99\pm0.28$ \\
CuCo & $64.43\pm1.57$ & $75.91\pm0.55$ & $79.24\pm0.56$ & $79.20\pm1.12$ & $72.30\pm0.34$ & $71.98\pm1.23$ & $51.32\pm1.89$ & $88.60\pm0.55$ & $56.49\pm0.19$ \\
AD-GCL & $65.23\pm1.34$
  & $75.04\pm0.48$ & $75.86\pm0.62$ & $75.73\pm0.51$ & $74.89\pm0.90$ & $71.49\pm0.98$ & $52.34\pm1.29$
  & \boldmath{}\textbf{$92.35\pm0.79$}\unboldmath{} & $56.24\pm0.43$ \\
\midrule
\textbf{nmrGCL} & \boldmath{}\textbf{$68.80\pm2.12$}\unboldmath{} & \boldmath{}\textbf{$78.01\pm1.23$}\unboldmath{} & \boldmath{}\textbf{$81.04\pm1.85$}\unboldmath{} & \boldmath{}\textbf{$80.61\pm1.25$}\unboldmath{} & \boldmath{}\textbf{$75.04\pm0.98$}\unboldmath{} & \boldmath{}\textbf{$73.83\pm1.76$}\unboldmath{} & \boldmath{}\textbf{$54.25\pm2.13$}\unboldmath{} & $90.79\pm1.23$ & \boldmath{}\textbf{$57.58\pm1.05$}\unboldmath{} \\
\bottomrule
\end{tabular}%
}
% \vspace{-10pt}
%   \caption{Unsupervised graph classification result comparison (\% with standard deviation) of \textbf{nmrGCL} and baselines. }
\label{tab:comp_result}
  \vspace{-15pt}
\end{table}%

\begin{table}[tb!]
  \centering
  
\renewcommand{\arraystretch}{0.8}
\caption{Unsupervised node classification performance comparison (\% with std).}
% \vspace{-10pt}
\scalebox{0.7}{
% Table generated by Excel2LaTeX from sheet 'node_classification'
\begin{tabular}{c|cc|cccccc}
\toprule
Data/Methods & GCN & GAT & DGI & MVGRL & MERIT & GCA & CCA-SSG & nmrGCL \\
\midrule
PubMed & 79.0 & $79.0\pm0.3$ & $77.3\pm0.6$ & $80.1\pm0.7$ & $80.1\pm0.4$ & $80.7\pm0.2$ & $81.6\pm0.4$ & \boldmath{}\textbf{$82.23\pm0.54$}\unboldmath{} \\
Coauthor-CS & $91.8\pm0.1$ & $90.5\pm0.7$ & $90.0\pm0.3$ & $92.1\pm0.1$ & $92.4\pm0.4$ & $92.95\pm0.12$ & \boldmath{}\textbf{$93.31\pm0.22$}\unboldmath{} & $91.67\pm0.33$ \\
Amazon-Photo & $87.3\pm1.0$ & $86.2\pm1.5$ & $83.1\pm0.5$ & $87.3\pm0.3$ & $87.4\pm0.2$ & $92.24\pm0.21$ & \boldmath{}\textbf{$93.14\pm0.14$}\unboldmath{} & $92.79\pm0.17$ \\
Computer & $86.5\pm0.5$ & $86.9\pm0.3$ & $83.9\pm0.5$ & $87.5\pm0.1$ & $86.9\pm0.5$ & $87.85\pm0.31$ & \boldmath{}\textbf{$88.74\pm0.28$}\unboldmath{} & $87.49\pm0.35$ \\
\bottomrule
\end{tabular}%
}
    % \vspace{-10pt}
    %   \caption{Node classification result comparison (\% with standard deviation) of \textbf{nmrGCL} and baselines.}
      \vspace{-10pt}
  \label{tab:node_classification}%
\end{table}%

\textbf{Ablation study.}
% To verify the superiority of our non-maximization removal operation for learning complementary embeddings, comprehensive ablation studies on graph classification are conducted. 
Ablation studies on graph classification are conducted to verify the superiority of our non-maximum removal operation for learning complementary embeddings. We design following 5 variants of the proposed \textbf{nmrGCL}: 
(1) \textbf{nmrGCL (w/o RO)}: delete the removal operation component. Note that this variant is the same as GraphCL which maximizes the similarity between embeddings of positive pairs. 
(2) \textbf{nmrGCL-bi}: bi-directional removal operation, i.e., for positive pairs, each embedding of augmented view conducts the non-max removal operation on the other. 
(3) \textbf{nmrGCL-rand}: for positive pairs, randomly mask some dimensions of embedding of the second augmented view with zeros, which is equivalent to applying a dropout function.
To make sure the ratio of masked dimensions in the variant is the same as in nmrGCL with the optimal hyper-parameter threshold $\delta$, as shown in Fig.~\ref{fig:threshold},  we implement the nmrGCL-rand by randomly shuffling $\mathbf{M}$ in Eq.~\ref{eq:mask}. 
% where the number of masked dimensions is consistent with the best in Fig~\ref{fig:threshold}. 
(4) \textbf{nmrGCL-non-min}: for positive pairs, the dimensions of the second embedding to be erased are the smallest dimensions of the first embedding instead of the largest ones.
(5) \textbf{nmrGCL-learn}: the removal mask is learned by randomly initialized learnable parameters with Sigmoid function.
% erase the dimensions in the embedding of the first augmented view that has the same \textit{minimum} dimension as the embedding of the second.

% (5) nmrGCL-var5: perform the removal operation after the projection header. 

Fig.~\ref{fig:ablation} compares nmrGCL and its variants, from which we make the following observations. 
First, the classification results decrease if the non-maximum removal operation is deleted, verifying the efficacy of leaning the complementary embedding. 
nmrGCL-rand improves the GraphCL but does not outperform the nmrGCL.
% nmrGCL-rand alleviates the overfitting problem in GraphCL to some extent and improves the classification accuracy, but nmrGCL-rand  is still not able to solve the problem of concentrating information in a few dimensions.  
% nmrGCL-rand 
% nmrGCL-bi outperforms nmrGCL in accuracy on only one dataset, suggesting that bidirectional removal operation prevents the encoder from learning enough representational information. 
nmrGCL-bi has the similar performance with nmrGCL.
The non-min removal operation can hardly improve and even hurts the model. 
nmrGCL-learn has different performance on different datasets. On datasets such as DD and IMDB-B which have a large edge-node ratio (715/284, 96/19), learnable mask strategy outperforms the nmrGCL. 
On NCI1, REDDIT-M datasets with small edge-node ratio (32/29, 594/508), the learnable mask strategy performs poorly.

\vspace{-3pt}
\begin{table}[tb!]
\hspace{-10pt}
\begin{minipage}[b]{.3\linewidth}
  \centering
 \includegraphics[width=0.8\linewidth]{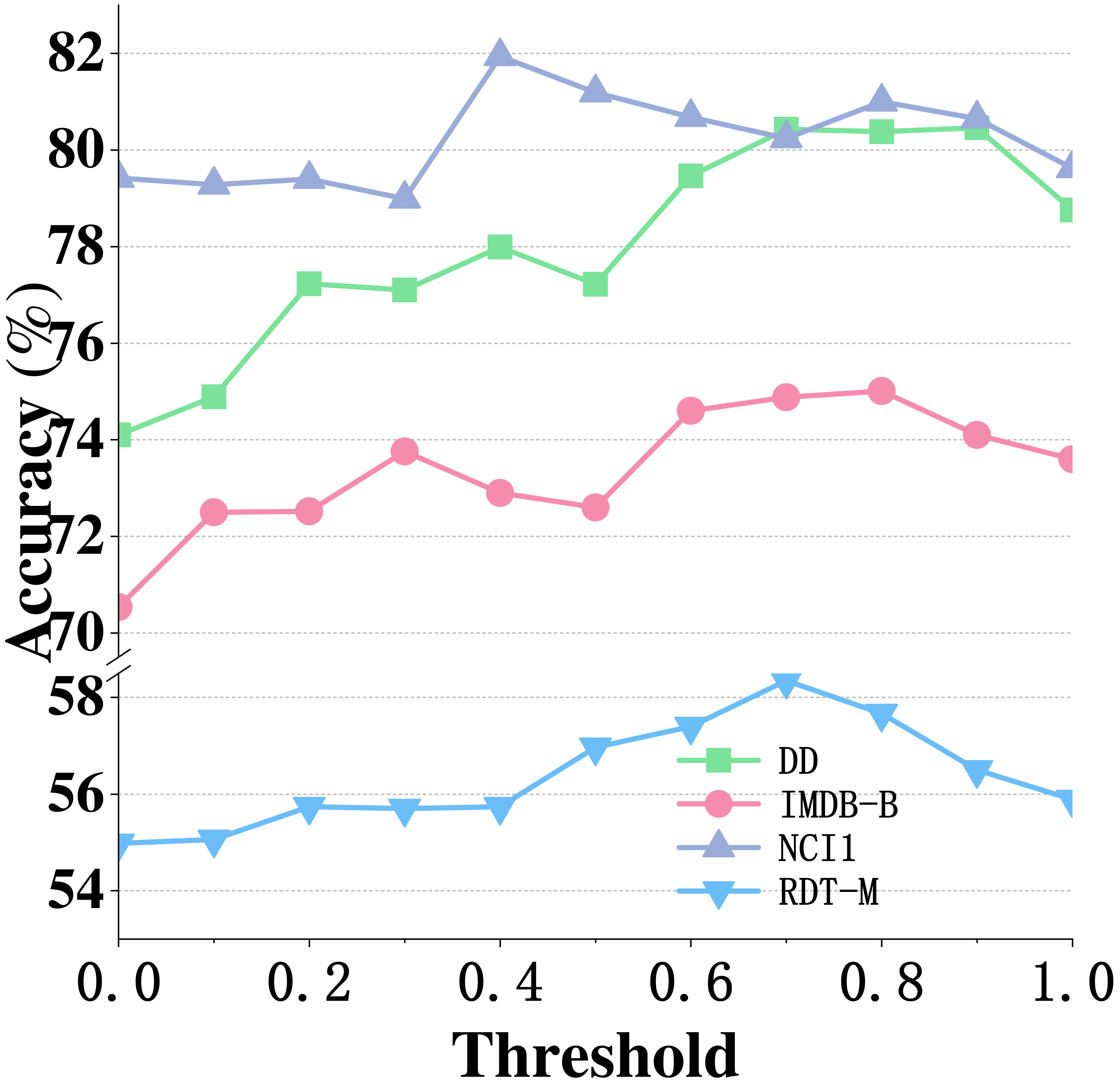}
% \vspace{-10pt}
  \captionsetup{width=0.8\linewidth}
%   \captionof{figure}{Threshold $\delta$ sensitivity study on bioinformatics and social network datasets.}
\captionof{figure}{Threshold $\delta$ sensitivity study.}
\label{fig:threshold}   
%  \vspace{-10pt}
%   \centerline{(a) Result 1}\medskip
\end{minipage}
\hspace{-17pt}
\begin{minipage}[b]{.3\linewidth}
  \centering
  \captionsetup{width=0.8\linewidth}
  \includegraphics[width=0.8\linewidth]{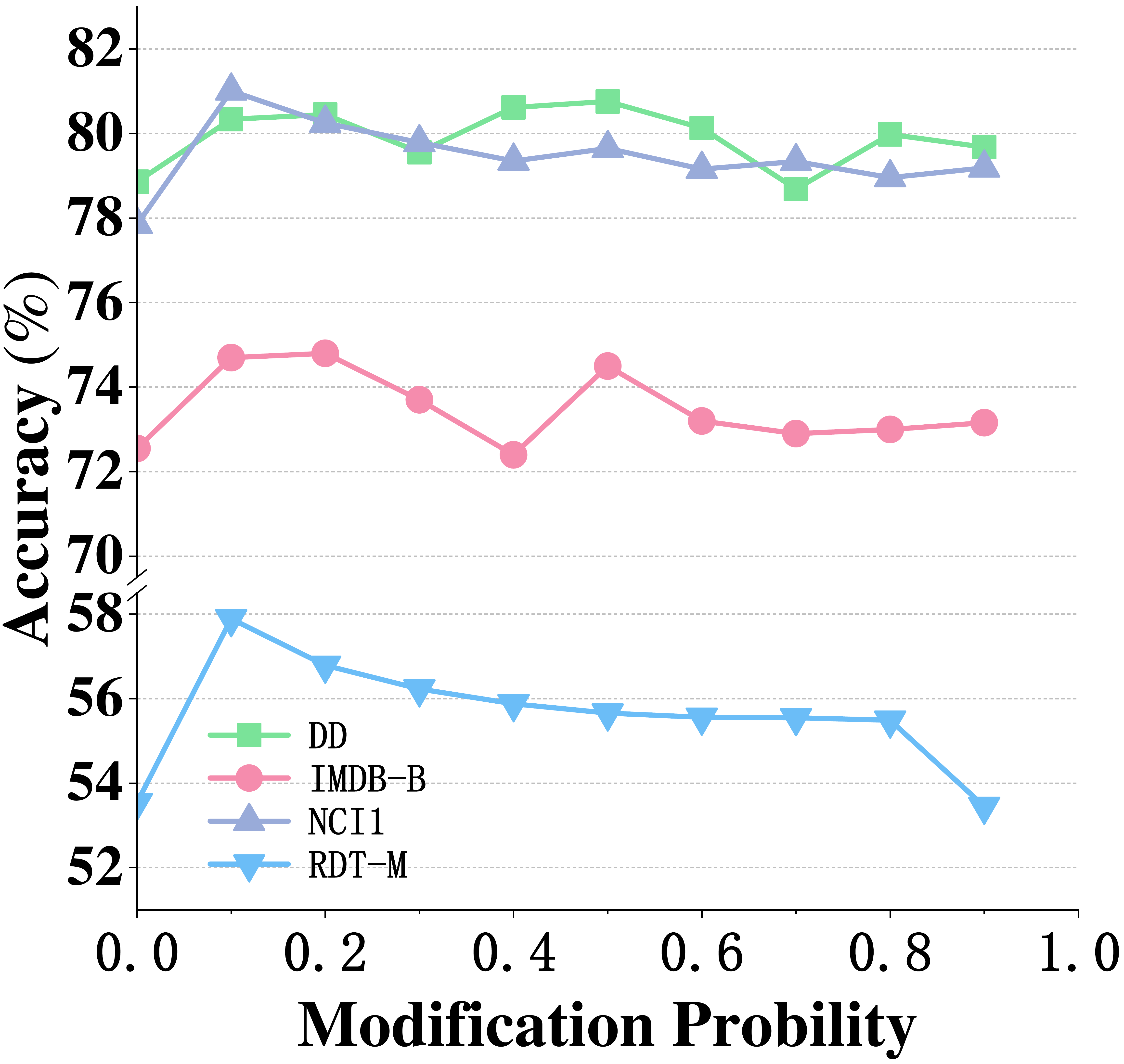}
    % \vspace{-10pt}
    % \captionof{figure}{ Effect of augmentation intensity, i.e., the modification probability of augmentation.}
    \captionof{figure}{ Effect of modification probability.}

\label{fig:aug}
% \vspace{-10pt}
\end{minipage}
\hspace{-5pt}
\begin{minipage}[b]{.44\linewidth}
% \vspace{-40pt}

\resizebox{\textwidth}{10mm}{
% Table generated by Excel2LaTeX from sheet 'transfer'
\begin{tabular}{c|cccc}
\toprule
% \midrule
{Pre-train} & \multicolumn{4}{c}{ZINC 2M}\\
\midrule
{Datasets} & BBBP & ToxCast & SIDER & ClinTox \\
\midrule
No-pre-train & $65.8\pm4.5$ & $63.4\pm0.6$ & $57.3\pm1.6$ & $58.0\pm4.4$ \\
EdgePred & $67.3\pm2.4$ & $64.1\pm0.6$ & $60.4\pm0.7$ & $64.1\pm3.7$ \\
AttrMasking & $64.3\pm2.8$ & \boldmath{} \textbf{$64.2\pm0.5$} \unboldmath{} & $61.0\pm0.7$ & $71.8\pm4.1$ \\
ContexPred & $68.0\pm2.0$ & $63.9\pm0.6$ & $60.9\pm0.6$ & $65.9\pm3.8$ \\
GraphCL & $69.68\pm0.67$ & $62.40\pm0.57$ & $60.53\pm0.88$ & $75.99\pm2.65$ \\
\midrule
nmrGCL & \boldmath{}\textbf{$70.34\pm1.21$}\unboldmath{} & $63.04\pm0.92$ & \boldmath{}\textbf{$61.45\pm1.63$}\unboldmath{} & \boldmath{}\textbf{$76.92\pm1.91$}\unboldmath{} \\
\bottomrule
\end{tabular}%
}
% \vspace{3pt}
% \vspace{10pt}
\caption{Graph classification accuracy (by mean and std) for transfer learning by ROC-AUC. The model is pre-trained on ZINC 2M dataset and transferred to other four datasets via fine-tuning. The settings follow \cite{pretrain}.}
% \vspace{20pt}
% \vspace{-20pt}
% \vspace{-10pt}
\end{minipage}
% \caption{Results of parameter sensitivity analysis. Left figure shows the effect of different removal threshold. Right figure shows the effect of intensity of augmentations.}
\vspace{-20pt}
\label{tab:transfer}
\end{table}

\subsection{Parameter Sensitivity and Transfer Learning Study}
\textbf{Sensitivity study w.r.t. the removal threshold $\delta$.} We vary the value of the crucial hyper-parameter threshold $\delta$ in the removal operation in Eq.~\ref{eq:mask} from 0.0 to 1.0 on four datasets, and the results are shown in Fig.~\ref{fig:threshold}. 
Unexpectedly even if the second embedding is completely removed when $\delta=0$, the randomly initialized untrained encoder still reaches a competitive performance, because the SVM can classify graphs directly based on node features. 
In general, we find that when the threshold $\delta$ is less than 0.7,  the classification accuracy grows with the increase of the threshold. The optimal value of $\delta$ is 0.6-0.8 for most datasets, with the exception of dataset NCI1, where the optimal value is 0.4.

\textbf{Sensitivity study w.r.t. intensity of augmentations.} 
Since the sparsity of augmented views affects the common prominent dimensions of both embeddings, we vary the ratio of nodes, edges or features discarded in graph augmentation including \textit{NodeDrop}, \textit{EdgeDrop} and \textit{FeatureMasking} on four datasets. 
From the results in Figure~\ref{fig:aug}, we observe that classification performance degenerates as the intensity of augmentation grows overly high. The optimal modification probability for most datasets is $0.1$ to $0.3$. These results are in record with the observation that graph-data are sparse and hard to be recovered after discarding information.

\textbf{Transfer learning study.} We conduct experiments on four large-scale datasets to evaluate the transferability in predicting the molecular property.
The encoder is  pre-trained  on the ZINC dataset without label and fine-tuned on other datasets, where all settings follow \cite{pretrain}. 
We select baselines including no-pre-trained GIN, GraphCL \cite{graphcl} and strategies used in \cite{pretrain} including EdgePred, AttrMasking and ContextPred. 
The result is shown in Tab.~3 with mean and standard deviation of ROC-AUC score for five trials.  
nmrGCL achieves the best performance on three of four datasets and outperforms GraphCL on all datasets. Detailed setup  of transfer learning is in Appendix~\ref{app:exp_config}.
% \vspace{-5pt}
\section{Conclusion and Broader Impact}
\vspace{-5pt}
% \subsection{Visulization}
% \subsection{Unsupervised Representation Learning}
% This paper investigates the gaps between contrastive loss and downstream tasks (graph/node classification) performance and proposes the graph complementary contrastive learning in a self-supervised manner named nmrGCL. 
% This paper investigates the gaps between contrastive loss and downstream tasks (graph/node classification) performance and proposes the graph complementary contrastive learning in a self-supervised manner named nmrGCL.
We have pointed out and theoretically analyzed a problem called dimensional collapse in graph contrastive learning (GCL), where information of embeddings concentrates on parts of dimensions.
We identify that graph pooling and convolution layers of GNNs specifically cause the dimensional collapse in GCL. 
% Graph contrastive learning suffers from dimensional collapse 
To alleviate the dimensional collapse, we propose the graph complementary contrastive learning named nmrGCL. 
% In nmrGCL, the first embedding of positive pairs discovers the most significant dimensions and erases these dimensions of the second embedding which is considered as the complement of the first. 
For each positive pair, nmrGCL identifies ``prominent'' dimensions in embedding of the first augmented view and erases these dimensions in the second, which is considered as the complement of the first.
The complementary embedding helps the encoder learn neglected information and enhance the distinguishability of the embedding. 
Experiments show that nmrGCL significantly outperforms the state-of-the-art SSL graph classification methods. We did not identify any technical limitation, nor potential negative impact of our methods to the society.

%The mainstream of graph networks are lightweights (usually have shallow layers, \yan{perhaps restricted to} the training speed and storage). 
\clearpage
\bibliography{neurips_2021}
\bibliographystyle{abbrv}

\newpage
\appendix
\section{More Related Works}
\textbf{Graph neural networks.}
Graph Neural Networks (GNNs) have attracted growing attention for analyzing graph-structured data in recent years.
Generally, GNNs are categorized into spatial-domain and spectral-domain approaches. Based on the spectral graph theory, \cite{spectral} first defines the graph convolution in the spectral domain through the eigen-decomposition of the graph Laplacian, defectively causing high computational cost.  
Graph Convolution Network \cite{gcn} utilizes the 1-st approximation of the Chebyshev expansion to simplify the calculation. 
% Following this, A series of spectral-based graph convolution  \cite{survey} has been carried out. 
Spatial-based approaches follow a message passing scheme \cite{mixhop}, where each node collects the information from its neighbors iteratively. 
GraphSAGE \cite{graphsage} aggregates the information from randomly sampled neighborhoods to scale to large graphs.
GAT \cite{gat} introduces the attention mechanism to assign scores for each node pair. 
% Nevertheless these methods are trained in a supervised manner, our method is designed for self-supervised graph representation learning. 
GIN \cite{gin} generalizes the Weisfeiler-Lehman test and reaches the most expressive power among GNNs. 
% AirGNN \cite{airgnn} utilized adaptive residual to improve the robustness.

\textbf{Comparison to DirectCLR\cite{dimensional_collapse}.} 
The theoretical analysis in DirectCLR does not apply to the graph domain due to the different mechanisms in GNN and CNN. DirectCLR attributes the dimensional collapse in vision field to two reasons: too strong augmentation on images and the interplay between adjacent fully connected layers.  
In comparison, our paper concentrates on the graph field and analyze that two specific component in GNN cause the dimensional collapse in the GCL, i.e., the graph pooling layers smooth the variance between embeddings of positive pair and the product of multi-layer graph convolutions has the tendency to be low-rank. 
In terms of the algorithms, our nmrGCL removes the prominent dimensions in one of positive pairs in pre-text tasks while the DirectCLR simply removes the projection head. 
Removal of projection head cannot improve the downstream classification accuracy in the graph domain since it cannot address the alignment between embeddings of positive pair. 
Our methods may be extended to visual contrastive learning since the implicit regularization exists in deep neural network.

% that embeddings of positive pair are highly correlated since variance caused by augmentation is smoothed by permutation-invariant pooling layer. 
% The alignment of embeddings makes the 
% Then we analyze that 

\section{Proofs in Sec.~\ref{sec:dimensional_collapse}}
\label{app:proof}
% \subsection{Proof of Lemma~\}
\subsection{Derivation of Gradient of InfoNCE w.r.t Embeddings}
\label{app:proof_g_z}
The InfoNCE loss for the first augmented view of $i$-th graph is given by: 
\begin{equation}
 \mathcal{L}_{i}^{(1)} = -\log\frac{\exp{(\langle \mathbf{z}_i^{(1)}, \mathbf{z}_i^{(2)}\rangle/\tau)}}{\exp{(\langle \mathbf{z}_i^{(1)}, \mathbf{z}_i^{(2)}\rangle/\tau)} + \sum_{l\in\{1,2\}, j\in\{1,\cdots ,N\}, j\neq i}\exp{((\langle \mathbf{z}_i^{(1)}, \mathbf{z}_j^{(l)}\rangle/\tau))}}, 
%  \label{eq:infonce}
\end{equation}
The gradient of InfoNCE w.r.t positive pairs embeddings can be derived: 
\begin{equation*}
\begin{aligned}
-&\frac{\partial \mathcal{L}_i^{(1)}}{\partial \mathbf{z}_i^{(1)} }\\
% -\nabla_{\mathbf{v}}  \mathcal{L}_{NCE}
=&\frac{\mathbf{z}_i^{(2)}}{\tau}-
\frac{\exp \left(\left\langle  \mathbf{z}_i^{(1)}, \mathbf{z}_i^{(2)}\right\rangle / \tau\right)}{\exp{\left(\left\langle  \mathbf{z}_i^{(1)}, \mathbf{z}_i^{(2)}\right\rangle / \tau\right)} + \sum_{l\in\{1,2\}, j\in\{1,\cdots ,N\}, j\neq i}\exp{((\langle \mathbf{z}_i^{(1)}, \mathbf{z}_j^{(l)}\rangle/\tau)) }}  \cdot \frac{  \mathbf{z}_i^{(2)}}{\tau}-\\
&\frac{\sum_{l\in\{1,2\}, j\in\{1,\cdots ,N\}, j\neq i} \exp \left(\left\langle  \mathbf{z}_i^{(1)},   \mathbf{z}_j^{(l)}\right\rangle / \tau\right)\cdot \mathbf{z}_j^{(l)} / \tau}{\exp{\left(\left\langle  \mathbf{z}_i^{(1)}, \mathbf{z}_i^{(2)}\right\rangle / \tau\right)} + \sum_{l\in\{1,2\}, j\in\{1,\cdots ,N\}, j\neq i}\exp{((\langle \mathbf{z}_i^{(1)}, \mathbf{z}_j^{(l)}\rangle/\tau)) }}\\
=&\frac{1}{\tau}
\left(\left(1- \frac{\exp(\langle \mathbf{z}_i^{(1)} , \mathbf{z}_i^{(2)} \rangle)}{p_i}\right) \cdot \mathbf{z}^{(2)}_i- \sum_{l\in\{1,2\}, j\in\{1,2,\cdots,N\},j\neq i}\left(\frac{\exp(\langle \mathbf{z}_i^{(1)}, \mathbf{z}_j^{(l)} \rangle)}{p_i}\cdot \mathbf{z}^{(l)}_j\right)\right)\\
% \end{aligned}
% \end{equation*}
% \begin{equation*}
% \begin{aligned}
&\\
-&\frac{\partial \mathcal{L}_i^{(1)}}{\partial \mathbf{z}_i^{(2)}}\\
=& \frac{1}{\tau}  \mathbf{z}_i^{(1)}-\frac{\exp \left(\left\langle  \mathbf{z}_i^{(1)}, \mathbf{z}_i^{(2)}\right\rangle / \tau\right)}{\exp{(\langle \mathbf{z}_i^{(1)}, \mathbf{z}_i^{(2)}\rangle/\tau)} + \sum_{l\in\{1,2\}, j\in\{1,\cdots ,N\}, j\neq i}\exp{((\langle \mathbf{z}_i^{(1)}, \mathbf{z}_j^{(l)}\rangle/\tau))}} \cdot \frac{ \mathbf{z}_i^{(1)}}{\tau}\\
=& \frac{1}{ \tau}\left(1- \frac{\exp(\langle \mathbf{z}_i^{(1)} , \mathbf{z}_i^{(2)} \rangle)}{p_i} \right)\cdot \mathbf{z}_i^{(1)}
\end{aligned}
\end{equation*}
where 
$$
p_i = \sum_{j=1, j\neq i}^N\sum_{l\in\{1,2\}}\exp{((\langle \mathbf{z}_i^{(1)}, \mathbf{z}_j^{(l)}\rangle/\tau))}.
% \end{equation}
$$
% \begin{equation}
% \begin{aligned}
% -\nabla_{\mathbf{v}^{-}}  \mathcal{L}_{NCE} &=\frac{1}{Y} \exp \left(\left\langle \mathbf{u},  \mathbf{v}^{-}\right\rangle / \tau\right) \cdot \frac{ \mathbf{u}}{\tau} \\
% &=\frac{ {1-{p}_{\mathbf{z}_i^{(2)}}}}{\tau} \cdot \frac{\exp \left(\left\langle \mathbf{u},  \mathbf{v}^{-}\right\rangle / \tau\right)}{U} \cdot \mathbf{u}
% \end{aligned}
% \end{equation}
The gradient of InfoNCE for the other view can be simply derived due to the symmetry. 
\subsection{Proof of Lemma~\ref{lemma:common_nodes}}
\label{app:proof_common_nodes}
\textbf{Lemma~\ref{lemma:common_nodes}} 
Given a graph $G$, two augmented views are generated by randomly discarding $p$-proportion nodes. The expected number of nodes shared by the two augmented graph is $(1-p)^2\cdot N$
\begin{proof}
Let $V$ be the set of nodes in $G$. and let $V^{(1)}$ and $V^{(2)}$ be sets of nodes in two augmented graphs. 
$V^{(1)}$ and $V^{(2)}$ are independent permutation of $V$, where $|V^{(1)}| = |V^{(2)}| = (1-p)|V|$. 
The probability that any node $n$ is in both augmented graph is $(1-p)^2$. 
By linearity of expectation, since the total number of nodes is $N$, the expectation of nodes in both augmented graphs is $(1-p)^2 \cdot N$, i.e., $\mathbb{E}[|V^{(1)} \cap V^{(2)}|] =(1-p)^2\cdot N$
\end{proof}
\subsection{Proof of Theorem~\ref{theorem:similarity}}
\label{app:proof_similarity}
\textbf{Theorem~\ref{theorem:similarity} }Given a cycle graph $G=(V,E)$ and node drop portion $p$, embeddings of two augmented views indexed by $1$ and $2$ are obtained by Eq.~\ref{eq:emb}. Then we have the lower bound for expectation of cosine similarity:
% \vspace{-10pt}
\begin{equation}
    \mathbb{E}[\texttt{sim}(\mathbf{z}^{(1)}, \mathbf{z}^{(2)})] > \frac{n+2}{(\sqrt{n}+1)^2}
\end{equation}
% \vspace{-5pt}
where $n$ is the ratio of common and unique nodes in two augmented views with expectation $\mathbb{E}[n]=(1-p)/p$ discussed in Lemma~\ref{lemma:common_nodes}.
\begin{proof}
A cycle graph $G=(V,E)$ is a simple graph such that $|V| = |E| =N (N>3)$. Thus, each node is connected exactly to two nodes. The feature represented by a one-hot vector of each node is $\mathbf{x}_i\in \mathbb{R}^F$. 
Two augmented graphs $G^{(1)} = (V^{(1)}, E^{(1)}), G^{(2)}=(V^{(2)}, E^{(2)})$ are generated by randomly dropping $p$ portion of nodes, $|V^{(1)}| = |V^{(2)}| = p\cdot N$ .
Denote the ratio of common and unique nodes in two augmented views with $n$:
$$
n := \frac{|V^{(1)} \cap V^{(2)}|}{|V^{(1)} - (V^{(1)} \cap V^{(2)})|} =\frac{|V^{(1)} \cap V^{(2)}|}{|V^{(2)} - (V^{(1)} \cap V^{(2)})|} 
$$
The expectation of $n$ can be derived with Lemma~\ref{lemma:common_nodes}:
\begin{equation}
\mathbb{E}[n] = \frac{(1-p)^2\cdot N}{(1-p)\cdot N - (1-p)^2\cdot N} = \frac{1-p}{p} \label{eq:E_n}
\end{equation}
% Without loss of generality, suppose that the first $x_1, x_2, \cdots, x_{c}$ are commonly retained in both augmented graphs.
Denote the set of commonly retained nodes in both augmented graphs as $V_c$, the set of uniquely retained nodes in each augmented nodes as $V_u^{(1)}, V_u^{(2)}$ for short. 
% Without loss of generality, assume that the nodes in $V_c, V_u^{(1)}, V_u^{(2)}$ are sequentially consecutive, i.e., $V_c=\{v_1, v_2, \cdots v_{|V_c|}\}, V_u^{(1)}=\{v_{|V_c|+1}, \cdots v_{p\cdot N}\}, V_u^{(2)}= \{v_{N-|V_c|},\cdots, v_{N}\}$. 
Without loss of generality, the nodes in these sets are selected sequentially and consecutively as follows: $V_c=\{v_1, v_ 2, \cdots v_{|V_c|}\}, V_u^{(1)}=\{v_{|V_c|+1}, v_{|V_c|+2}, \cdots, v_{p\cdot N}\}, V_u^{(2)}= \{v_{N-|V_c|},\cdots,v_{N-1}, v_{N}\}$. Given the single layer graph convolution defined in Eq.~\ref{eq:emb}, the aggregated node features vectors of two augmented graph become: 
\begin{align*}
    X^{(1)} &= \{ (\mathbf{x}_1 + \mathbf{x}_{2}/\sqrt{2})\cdot \mathbf{W},  \cdots, \frac{\mathbf{x}_{|V_c|-1}+\mathbf{x}_{|V_c|}+\mathbf{x}_{|V_c|+1}}{2} \cdot \mathbf{W}, \cdots, (\mathbf{x}_{p\cdot N-1} /\sqrt{2}+\mathbf{x}_{p\cdot N}\cdot) \mathbf{W} \}\\ 
    X^{(2)} &= \{\frac{\mathbf{x}_N +\mathbf{x}_1 + \mathbf{x}_{2}}{2}  \cdot \mathbf{W}, \cdots,  (\mathbf{x}_{|V_c|-1}/\sqrt{2} + \mathbf{x}_{|V_c|})\cdot \mathbf{W},  \cdots, \frac{\mathbf{x}_{N-1} +\mathbf{x}_N + \mathbf{x}_{1}}{2} \cdot \mathbf{W}\}
\end{align*}
Applying the average pooling (readout) layer  on node embeddings, the graph-level embeddings can be obtained with:
\begin{align*}
    \mathbf{z}^{(1)} &= \frac{1}{p\cdot N}\left ( \sum_{i=1}^{|V_c|}\frac32\mathbf{x}_i + \sum_{i=|V_c|+1}^{p\cdot N} \frac{2}{3}\mathbf{x}_i - \frac{1-\sqrt{2}}{2}(\mathbf{x}_1+\mathbf{x}_2 + x_{p\cdot N} + x_{p\cdot N -1})\right)\\
    \mathbf{z}^{(2)} &= \frac{1}{p\cdot N}\left ( \sum_{i=1}^{|V_c|}\frac32\mathbf{x}_i + \sum_{i=N-|V_c|}^{N} \frac23\mathbf{x}_i - \frac{1-\sqrt{2}}{2}(\mathbf{x}_{|V_c|}+\mathbf{x}_{|V_c|-1} + \mathbf{x}_{N-|V_c|} + \mathbf{x}_{p\cdot N -|V_c|+1} \right)
\end{align*}
The expression for cycle graph embeddings can be generalized to regular graphs. 
Since the number of nodes $N$ can be very large in real datasets, thus we can have the approximation:
\begin{equation}
     \mathbf{z}^{(1)} \approx \frac{3}{2 p\cdot N}\left ( \underbrace{\sum_{i=1}^{|V_c|}\mathbf{x}_i}_{:=\mathbf{S}_c} + \underbrace{\sum_{i=|V_c|+1}^{p\cdot N} \mathbf{\mathbf{x}_i}}_{:=\mathbf{S}_1}  \right)  \quad 
     \mathbf{z}^{(2)} \approx \frac{3}{2 p\cdot N}\left ( \underbrace{\sum_{i=1}^{|V_c|}\mathbf{x}_i}_{:=\mathbf{S_c}}+ \underbrace{\sum_{i=N-|V_c|}^{N} \mathbf{\mathbf{x}_i}}_{:\mathbf{S}_2} \right)
\end{equation}
Note that the embedding can be divided into two parts: the aggregation of common nodes features and the aggregation of unique nodes features. 

% Before we analyze the similarity between two embeddings, we make the following assumption: 
% Assume that the one-hot vector feature of each node contains $F/2$ of ones, i.e., $\Vert\mathbf{x}_1\Vert = \Vert\mathbf{x}_2\Vert= \cdots =\Vert\mathbf{x}_N\Vert = \sqrt{F/2}$. 
Assume that each element in one-hot encoding feature of each node follows a Bernoulli distribution, $\forall i,j,  x_i[j] \sim \text{Bern}(q)$. Then we have:
\begin{align*}
    \mathbb{E}(\Vert\mathbf{S}_c\Vert) =  \sqrt{V_c \cdot F\cdot q}, \quad \mathbb{E}(\Vert \mathbf{S}_1 \Vert)=\mathbb{E}(\Vert \mathbf{S}_1 \Vert) =\sqrt{(p\cdot N -|V_c|)\cdot F \cdot q}
\end{align*}
Then the cosine similarity between two embeddings becomes: 
\begin{align*}
    \texttt{sim}(\mathbf{z}^{(1)}, \mathbf{z}^{(2)}) &=  \frac{\langle \mathbf{z}^{(1)}, \mathbf{z}^{(2)} \rangle}{ \Vert \mathbf{z}^{(1)}\Vert  \cdot \Vert  \mathbf{z}^{(2)}\Vert }\\
    &= \frac{ \langle (\mathbf{S}_c + \mathbf{S}_1), (\mathbf{S}_c + \mathbf{S}_2) \rangle }{\Vert\mathbf{S}_c + \mathbf{S}_1  \Vert  \cdot \Vert \mathbf{S}_c + \mathbf{S}_2\Vert  }\\
    &\geq \frac{\langle (\mathbf{S}_c + \mathbf{S}_1), (\mathbf{S}_c + \mathbf{S}_2) \rangle }{ (\Vert \mathbf{S}_c\Vert+\Vert \mathbf{S}_1\Vert) \cdot (\Vert \mathbf{S}_c\Vert+\Vert \mathbf{S}_2\Vert) }\\
    &= \frac{\langle \mathbf{S}_c, \mathbf{S}_c \rangle +\langle \mathbf{S}_c, \mathbf{S}_1 \rangle + \langle \mathbf{S}_c, \mathbf{S}_2 \rangle  + \langle \mathbf{S}_1, \mathbf{S}_2 \rangle }{ \Vert \mathbf{S}_c\Vert^2 +\Vert \mathbf{S}_c\Vert \cdot \Vert \mathbf{S}_1\Vert+\Vert \mathbf{S}_c\Vert \cdot \Vert \mathbf{S}_2\Vert+\Vert \mathbf{S}_1\Vert \cdot \Vert \mathbf{S}_2\Vert}
\end{align*}
% Since $\Vert \mathbf{x}_1\Vert =\Vert \mathbf{x}_1\Vert =\cdots \Vert \mathbf{x}_N \Vert= \sqrt{F/2}$, we have following inequalities: 
Additionally, the cosine similarity reaches the minimum when $(\mathbf{S}_c-\mathbf{S}_1)\perp\mathbf{S}_1$ and $(\mathbf{S}_c-\mathbf{S}_2)\perp \mathbf{S}_2$.
\begin{align*}
    \texttt{sim}(\mathbf{z}^{(1)}, \mathbf{z}^{(2)}) &\geq \frac{ \Vert \mathbf{S}_c\Vert^2 +  \Vert \mathbf{S}_1 \Vert^2 +  \Vert \mathbf{S}_2 \Vert^2  + \langle \mathbf{S}_1, \mathbf{S}_2 \rangle }{ \Vert \mathbf{S}_c\Vert^2 +\Vert \mathbf{S}_c\Vert \cdot \Vert \mathbf{S}_1\Vert+\Vert \mathbf{S}_c\Vert \cdot \Vert \mathbf{S}_2\Vert+\Vert \mathbf{S}_1\Vert \cdot \Vert \mathbf{S}_2\Vert}\\
    &\geq \frac{ \frac{\Vert \mathbf{S}_c\Vert^2}{\Vert \mathbf{S}_1 \Vert^2} +  2}{ \frac{\Vert \mathbf{S}_c\Vert^2}{\Vert \mathbf{S}_1 \Vert^2} +\frac{\Vert \mathbf{S}_c\Vert}{\Vert \mathbf{S}_1 \Vert}+ \frac{\Vert \mathbf{S}_c\Vert}{\Vert \mathbf{S}_2 \Vert}+1}\\
    % &= \frac{n^2+2}{(n+1)^2}
\end{align*} 
Thus, the expectation of similarity is bounded by:
\begin{align}
    \mathbb{E}[\texttt{sim}(\mathbf{z}^{(1)}, \mathbf{z}^{(2)})] &\geq \frac{\frac{V\cdot F \cdot q}{(p\cdot N -|V_c|)\cdot F\cdot q}+2}{\frac{V\cdot F \cdot q}{(p\cdot N -|V_c|)\cdot F \cdot q} +  2 \cdot \sqrt{\frac{V\cdot F \cdot q}{(p\cdot N -|V_c|)\cdot F \cdot q}}+1} \\
    &= \frac{n+2}{n+2\sqrt{n}+1}
\end{align}
where $n$ is the ratio of common nodes and unique nodes in two augmented views with expectation $\mathbb{E}[n]=(1-p)/p$ in Eq.~\ref{eq:E_n}.
% where $n$ is bounded by
% $$
% \frac{\sqrt{|V_c|}}{p\cdot N -|V_c|}  \leq n\leq \frac{|V_c|}{\sqrt{p\cdot N - |V_c|}}, \quad \mathbb{E}[|V_c|] = (1-p)^2\cdot N
% $$
\end{proof}

\subsection{Proof of Lemma~\ref{lemma:g_sigma}}
\label{app:proof_g_sigma_lemma}
\textbf{Lemma~\ref{lemma:g_sigma}}
The derivative of singular values of the matrix product $\mathbf{W}$ w.r.t time are:
\begin{equation}
    \dot{\sigma_m}(t) = \mathbf{u}_m^\top(t) \dot{\mathbf{W}}(t) \mathbf{v} (t),
\end{equation}
where $\sigma_m$ is the $m$-th singular value of $\mathbf{W}$, $\mathbf{u}_m$ and $\mathbf{v}_m$ is the $m$-th column of $\mathbf{U}$ and $\mathbf{V}$, respectively.

\begin{proof}
Given a matrix $\mathbf{W}$ with singular value decomposition $\mathbf{W}= \mathbf{U}\mathbf{\Sigma}\mathbf{V}^\top$. 
Differentiate the SVD of $\mathbf{W}$ w.r.t time:
$$
\dot{\mathbf{W}}(t) =\dot{ \mathbf{U}}(t)\mathbf{\Sigma}(t)\mathbf{V}^\top(t) + \mathbf{U}(t)\dot{\mathbf{\Sigma}}(t)\mathbf{V}^\top(t)+ \mathbf{U}(t)\mathbf{\Sigma}(t)\dot{\mathbf{V}}^\top(t),
$$
Since $\mathbf{U}(t)$ and $\mathbf{V}(t)$ have orthonormal columns, multiply $\mathbf{U}^\top(t)$ from the left and $\mathbf{V}(t)$ from the right and have:
$$
\mathbf{U}^\top(t)\dot{\mathbf{W}}(t)\mathbf{V}(t) = \mathbf{U}^\top(t)\dot{ \mathbf{U}}(t)\mathbf{\Sigma}(t) + \dot{\mathbf{\Sigma}}(t)+ \mathbf{\Sigma}(t)\dot{\mathbf{V}}^\top(t)\mathbf{V}(t),
$$
Since $\mathbf{\Sigma}(t)$ is a diagonal matrix where $m$-th diagonal entry equals the $m$-th singular value $\sigma_m$, we focus on the diagonal singular values:
\begin{equation}
\mathbf{u_m}^\top(t)\dot{\mathbf{W}}\mathbf{v}_m(t) = \langle  \mathbf{u_m}(t), \dot{\mathbf{u_m}}(t) \rangle\sigma_m(t)+ \dot{\sigma}_m(t) + \sigma_m(t)\langle \dot{\mathbf{v}}_m(t), \mathbf{v}_r(t)\rangle 
\end{equation}
Since $\mathbf{u}_m$ and $\mathbf{v}_m$ are unit-norm bases, we have $ \langle  \mathbf{u_m}(t), \dot{\mathbf{u_m}}(t) \rangle = \frac{1}{2}\frac{d}{dt}\Vert  \mathbf{u_m}(t) \Vert^2_2 = 0$ and $\langle \dot{\mathbf{v}}_m(t), \mathbf{v}_r(t)\rangle=0$ similarly. 
Thus the equation becomes:
\begin{equation}
\dot{\sigma_m}(t) = \mathbf{u}_m^\top(t) \dot{\mathbf{W}}(t) \mathbf{v} (t)
\end{equation}
where $\mathbf{u}_m$ and $\mathbf{v}_m$ are $m$-th columns of matrices $\mathbf{U}$ and $\mathbf{V}$.
\end{proof}
\subsection{Proof of Theorem~\ref{theorem:g_sigma}}
\label{app:proof_g_sigma}
\textbf{Theorem 2}
The singular values of the product matrix $W(t)$ evolve by:
\begin{equation}
\dot{\sigma_m}(t) = -K\cdot (\sigma_m^2 (t))^{\frac{K-1}{K}} \cdot \mathbf{u}_m^\top(t) \cdot \mathbf{C_z} \cdot\mathbf{v}_m^{(t)},
\end{equation}
where $\mathbf{C_\mathbf{z}} = \sum_i (\mathbf{g}_{\mathbf{z}_i^{(1)}} \widetilde{\mathbf{A}}^K\mathbf{X}_i^{(1)} + \mathbf{g}_{\mathbf{z}_i^{(2)}}\widetilde{\mathbf{A}}^K\mathbf{X}_i^{(2)}) $, and $g_{\mathbf{z}_i^{(k)}}$ is the gradient over $\mathbf{z}_i^{(k)}$ in Eq.~\ref{eq:g_z}. 
\begin{proof}
We first adopt the following equation from Theorem 1 in \cite{implicit_regularization}:
\begin{equation*}
    \dot{\mathbf{W}(t)} = \sum_{j=1}^N \left[\mathbf{W}^\top(t)\mathbf{W}(t)\right]^\frac{K-j}{K}\cdot\frac{\partial \mathcal{L}(\mathbf{W}(t))}{\partial\mathbf{W}} \cdot \left[\mathbf{W}(t)\mathbf{W}^\top(t)\right]^\frac{j-1}{K}
\end{equation*}
where $[\cdot]^\alpha, \alpha\in\mathbb{R}^+ $ is the power operator on positive semi-definite matrices. 
Substitute the singular value decomposition of matrix in Eq.~\ref{eq:svd} into the above equation:
\begin{align*}
\dot{W}(t) =  &\sum_{j=1}^N \left[\mathbf{W}^\top(t)\mathbf{W}(t)\right]^\frac{K-j}{K}\cdot \nabla\mathcal{L}(\mathbf{W}(t)) \cdot \left[\mathbf{W}(t)\mathbf{W}^\top(t)\right]^\frac{j-1}{K}\\
= &-\mathbf{V}(t)(\mathbf{\Sigma}^2(t))^\frac{K-1}{K} \mathbf{V}^\top(t) \nabla\mathcal{L}(\mathbf{W}(t))\\\
&-\sum_{j=2}^{K-1}\mathbf{V}(t)(\mathbf{\Sigma}^2(t))^\frac{K-j}{K}\mathbf{V}^\top(t) \cdot    \nabla\mathcal{L}(\mathbf{W}(t))\cdot      \mathbf{U}(t)(\mathbf{\Sigma}^2(t))^\frac{j-1}{K}\mathbf{U}^\top (t)  \\
&-\nabla\mathcal{L}(\mathbf{W}(t))\mathbf{U}(t)(\mathbf{\Sigma}^2(t))^\frac{K-1}{K}\mathbf{U}^\top(t)
\end{align*}
Again, since  $\mathbf{V}$ and $\mathbf{V}$ consist of orthonormal columns, i.e., $ \forall i,j, i\neq j, \mathbf{u}_i^\top\mathbf{u}_j = \mathbf{v}_i^\top\mathbf{v}_j=0$. Thus, we multiply $\mathbf{v}^\top_m$ from left side and $\mathbf{u}_m$ from right side on both hands of equation: 
\begin{align*}
    \mathbf{v}^\top_m(t) \dot{\mathbf{W}}(t) \mathbf{u}_m(t)=  & -(\sigma_m^2(t))^\frac{K-1}{K}\mathbf{v}_m^\top \nabla\mathcal{L}(\mathbf{W}(t))\mathbf{u}_m(t)\\
    &- \sum_{j=2}^{N-1}(\sigma_m^2(t))^\frac{K-j}{K}\mathbf{v}^\top_m(t)\nabla\mathcal{L}(\mathbf{W}(t))\mathbf{u}_m (\sigma_m^2(t))^\frac{j-1}{K}\\
    &- \mathbf{v}_m^\top \nabla\mathcal{L}(\mathbf{W}(t)) \mathbf{u}_m(t) (\sigma^2_m)^\frac{K-1}{K}\\
    = & - K \cdot (\sigma_m^2(t))^\frac{K-1}{K} \mathbf{v}_m^\top \nabla \mathcal{L}(\mathbf{W}(t)) \mathbf{v}_m(t) 
\end{align*}
% Substituting the above equation into Eq.~\ref{eq:g_sigma}, we then have:
The gradient of loss on $W$ also can be obtained with the chain rule:
\begin{align*}
    \frac{d \mathcal{L}}{d W} = \sum_{a\in\{1,2\}} \sum_i \left(\frac{\partial \mathcal{L}}{\partial \mathbf{z}_i^{(a)}} \frac{\partial \mathbf{z}_i^{(a)}}{\partial \mathbf{W}} \right)
\end{align*}
Since $\mathbf{z}_i^{(a)}= \widetilde{A}^K\mathbf{X}^{(a)} \mathbf{W}$, we have $\frac{\partial \mathbf{z}_i^{(a)}}{\partial \mathbf{W}} = \widetilde{A}^K \mathbf{X}^{(a)}$. The gradients of InfoNCE loss w.r.t embedding vectors are derived in Eq.~\ref{eq:g_z} and here denoted as $g_{\mathbf{z}_i^{(a)}}$. 
Thus, finnaly we have the following expression for derivative of singular values:
\begin{equation}
    \dot{\sigma}_m(t) =  -K\cdot (\sigma_m^2 (t))^{\frac{K-1}{K}} \cdot \mathbf{u}_m^\top(t) \cdot \mathbf{C_z} \cdot\mathbf{v}_m^{(t)},
\end{equation}
where $\mathbf{C_\mathbf{z}} = \sum_i (\mathbf{g}_{\mathbf{z}_i^{(1)}} \widetilde{\mathbf{A}}^K\mathbf{X}_i^{(1)} + \mathbf{g}_{\mathbf{z}_i^{(2)}}\widetilde{\mathbf{A}}^K\mathbf{X}_i^{(2)}) $, and $g_{\mathbf{z}_i^{(k)}}$ is the gradient over $\mathbf{z}_i^{(k)}$ in Eq.~\ref{eq:g_z}. 
\end{proof}
\section{Details of Used Graph Augmentation Approaches}
\label{app:augmentation}
Given a graph $G = (V, E, \mathbf{X})$ where $V$ is the node set, $E$ is the edge set, and $\mathbf{X}$ is feature matrix.  The augmented graph view $\tilde{G}$ can be represented get with following methods:

\textbf{EdgeAdd} and \textbf{EdgeDrop} purtubs the adjacency matrix by randomly add or discard a portion of edges in graphs. Formally, the adjacency matrix of augmented graph is: 
$$
\tilde{\mathbf{A}} = M_d \odot \mathcal{A} + M_a\odot (1-\mathbf{A}), 
$$
where $M_d, M_a\in\{0,1\}^{N\times N}$ are dropping and adding matrix. 
% Each element of $M_1$ and $M_2$ is randomly sampled by Bernoulli distribution with pre-defined parameter $p$. 
Specifically, $M_d$ and $M_a$ first copy $\mathcal{A}$  and $(1-\mathbf{A})$ respectively and then randomly mask a portion of ones with zeros, where the ratio of dropped or added edges is a hyper-parameter $p$. 

\textbf{NodeDrop} perturbs the structure of the given graph by randomly discarding a portion of nodes with their features and connected edges. Since most GNN are utilized in transductive manner, we do not consider NodeAdd in this paper for node perturbations. Similar to EdgeDrop, each node in graphs has a probability of $p$ to be dropped. Then the adjacent edges in adjacency matrix  and features of dropped nodes  are also masked with zero. 

\textbf{NodeShuffling} perturbs the structure of the given graph by randomly shuffle a portion of nodes with their features and connected edges. Feature vectors of $p$-portion nodes are put into other positions while the adjacency matrix remains the same. 

\textbf{FeatureMasking} randomly masks a portion of features in all nodes. We first generate a mask $M_F\in \{0,1\}^F$ where each entry follows a Bernoulli distribution  with parameter $(1-p)$. Then the node feature matrix of augmented graph is:
$$
\tilde{\mathbf{X}} = [x_1\odot M_F, x_2\odot M_F, \cdots, x_n\odot M_F],
$$
where $\odot$ is the element-wise product. 
\textbf{FeatureDropout}  masks feature vectors of $p$-portion nodes with zeros.

\textbf{SubgraphSmapling} generates a subgraph with random walks with following steps. 
Start random work from a node. The walk has a probability which is proportional to edge weights to travel to neighbors and a hyper-parameter probability $p$ to go back to the start node. 
The augmented graph is the sub-graph induced by walked nodes in previous travels. 
Then each node in sub-graph is reordered by the sequence of appearance when traveling. 
% Node Drop 
% \label{app:augmentation}

\section{Training Algorithm}
\label{app:training_algoirthm}
\begin{algorithm}[htb!]
\caption{Training procedure of non-maximum removal for graph contrastive learning}
\label{alg:nmrGCL}
\textbf{Input}: Training set $\mathcal{G}
=\{G_j\}_{j=1}^{|\mathcal{G}|}
$, GNN encoder  $g_\theta(\cdot)$, augmentation distribution $\mathcal{T}$, threshold $\delta$, mask matrix $M$, batch size $B$\\
% \textbf{Parameter}: Optional list of parameters\\
\textbf{Output}:The pre-trained encoder $g_\theta(\cdot)$
\begin{algorithmic}[1] %[1] enables line numbers
\STATE Randomly initialize parameters $\theta$ of the GNN encoder and set all elements of $M$ to be 0.
\FOR{each mini-batch $\mathcal{B}$ sampled from $\mathcal{G}$}
\FOR{$k=1,2,\cdots, B$}
    \STATE Select two augmentation methods $t_1, t_2$ from $\mathcal{T}$
    \STATE $\tilde{G}_k^{(1)}\leftarrow t_1(G_k)$, $\tilde{G}_k^{(2)}\leftarrow t_2(G_k)$
    \STATE $r^{(1)}_k \leftarrow g_\theta(\tilde{G}_k{(1)})$, $r^{(2)}_k \leftarrow g_\theta(\tilde{G}_k^{(2)})$
    \STATE $M_{pq}=0$ if $|r_k^{(1)}|_{pq} > \delta$ where $p,q$ are indexes of elements in tensors. 
    \STATE $\hat{r}_k^{(2)} \leftarrow r_k^{(2)} \odot M $
    \STATE $z_{k,\pm}^{(1)} \leftarrow h(r_k^{(1)})$, 
$z_{k,+}^{(2)} \leftarrow h(\hat{r}_k^{2})$,  $z_{k,-}^{(2)} \leftarrow h(r_k^{2})$
    \STATE Computer the loss $\mathcal{L}$ via Eq.~\ref{eq:loss}.
    \STATE Update the parameters of $g_\theta(\cdot)$ and  $h(\cdot)$ with adam optimizer by mininizing $\mathcal{L}$.
\ENDFOR
\ENDFOR
\STATE \textbf{return} The GNN encoder $g_\theta(\cdot)$
\end{algorithmic}
\end{algorithm}
\section{Experiment Setup}
\label{app:exp}
\subsection{Software and Software infrastructures}
\label{app:software}
Our code is built based on \cite{graphcl}\footnote{https://github.com/Shen-Lab/GraphCL, MIT license}, \cite{zhu2021empirical}\footnote{https://github.com/PyGCL/PyGCL, Apache-2.0 license} for graph classification and \cite{dgcrl}\footnote{https://github.com/CRIPAC-DIG/GRACE, Apache-2.0 license} for node classification  with:
\begin{itemize}
    \item Software dependencies: Python 3.7.11, Pytorch 1.9.1, Pytorch-geometric 2.0.1, Numpy 1.20.1, scikit-learn 0.24.2, DGL-cuda10.1 0.7.1, scipy 1.6.2.
    \item CPU: Intel(R) Xeon(R) Gold 5222 CPU @ 3.80GHz
    \item GPU: 2 NVIDIA GeForce RTX 3090
    \item OS: Ubuntu 18.04.6 LTS
\end{itemize}
\subsection{Datasets}
\label{app:datasets}
\textbf{Unsupervised Graph Classification Datasets.} We first  adopt 4 bio-informatics datasets and 5 social network dataset from \cite{tudataset} with statistics shown in Tab.~\ref{tab:dataset}.
Four bio-informatics dataset including \textit{PCT-MR}, \textit{PROTEINCS}, \textit{NCI1} and \textit{DD} represent the molecules of proteins and other chemical compounds as graphs where nodes are amino acids or atoms and edges are chemical bonds. 
Graphs are divided into different classes according to their chemical classification or whether they have chemical reaction when encountering a particular substance. 
\textit{COLLAB} dataset is a scientific collaboration dataset which generate ego-networks of researchers in three field High Energy, Condensed Matter and Astro Physics. 
The label of each collaboration graph is the field that the author belongs to. 
\textit{IMDB-BINARY} and \textit{IMDB-MULTI} are movie collaboration dataset where the graphs represent movies,  nodes are actors/actresses, and edges represent existence of collaboration. 
The label of each graph is Action or Romance in \textit{IMDB-BINARY} and Comedy, Romance and Sci-Fi in \textit{IMDB-MULTI}.
\textit{REDDIT-BINARY} and \textit{REDDIT-MULTI} are social network datasets where graphs are on line discussion thread, nodes are users, and edges represent whether at least one node in each node pair respond to comments of another. 
\textit{REDDIT-BINARY} and \textit{REDDIT-MULTI} label each graph into 2 classes and 12 classes, respectively. 

\textbf{Unsupervised Node Classification Datasets.} We use 4 benchmark datasets including Pubmed, Coautho-CS, Amazon-Photo and Amazon-Computers \cite{shchur2018pitfalls} with statistics shown in Tab.~\ref{tab:dataset}. 
\textit{Pubmed} is a widely used node classification dataset containing one citation network, where nodes are papers and edges are citation relationships.
\textit{Coauthor-CS} are co-authorship graphs where nodes are authors and edges represents the existence of co-authorship in any paper. 
The labels of nodes are most active fields which each author study  on and the node features are keywords of each author's papers. 
\textit{Amazon-Photo} and \textit{Amazon-Computers} are two co-purchase networks where nodes are products and connectivity of nodes depends on frequency of co-purchase. 
Each node has a  feature vector according to product review and is labeled with categories.

\textbf{Graph Classification Datasets in Transfer Learning.} We use \textit{ZINC} datasets for pre-training and for 4 datasets \textit{BBBP}, \textit{ToxCast}, \textit{SIDER}, and \textit{CinTox}for downstream tasks. 
These datasets contain biological interactions and chemical molecules from \cite{pretrain} . 
\begin{table}[tb!]
    \centering
    \renewcommand{\arraystretch}{0.7}
    \scalebox{0.7}{
\begin{tabular}{c|c|c|c|c}
\toprule
Datasets & \# Graphs & Avg. Nodes & Avg. Edges & \# Classes \\
\midrule
\multicolumn{5}{c}{Unsupervised Graph Classification}\\
\midrule
PTC-MR & 344 & 14.29  & 14.69  & 2  \\
PROTEINS & 1113 & 39.06  & 72.82  & 2  \\
NCI1 & 4110 & 29.87  & 32.30  & 2  \\
DD & 1178 & 284.32  & 715.66  & 2  \\
\midrule
COLLAB & 5000 & 74.79  & 2457.78  & 3  \\
IMDB-BINARY & 1000 & 19.77  & 96.53  & 2  \\
IMDB-MULTI & 1500 & 13.00  & 65.94  & 3  \\
REDDIT-BINARY & 2000 & 429.63  & 497.75  & 2  \\
REDDIT-MULTI & 4999 & 508.52  & 594.87  & 5  \\
\midrule
\multicolumn{5}{c}{Unsupervised Node Classification}\\
\midrule
Pubmed & 1  & 19717 & 88651 & 3  \\
CS & 1  & 18333 & 327576 & 15  \\
Photo & 1  & 7650 & 287326 & 8  \\
Computers & 1  & 13752 & 574418 & 10  \\
\bottomrule
\end{tabular}%
}
% \vspace{-10pt}
    \caption{Statistics of the unsupervised graph  and node classification datasets.}
    % \vspace{-10pt}
    \label{tab:dataset}
\end{table}

\begin{table}[]
\renewcommand{\arraystretch}{0.7}  
\centering
\scalebox{0.8}{
\begin{tabular}{c|c|c|c|c}
\toprule
Datasets & \# Graphs & Avg. Nodes & Avg. Edges & \# Binary Tasks \\
\midrule
ZINC & 2e6 & 26.62 & 57.72 & Unlabeled  \\
BBBP & 2039 & 24.06 & 51.9 & 1  \\
ToxCast & 8576 & 18.78 & 38.52 & 617  \\
SIDER & 1427 & 33.64 & 70.71 & 27  \\
CinTox & 1477 & 26.15 & 55.76 & 2  \\
\bottomrule
\end{tabular}%
}
% \hspace{-10pt}
\caption{Statistics of graph classification datasets in transfer learning}
% \hspace{-10pt}
\label{tab:my_label}
\end{table}
\subsection{Experiment Configuration}
\label{app:exp_config}
\textbf{Unsupervised graph classification.} 
The model is evaluated following \cite{graphcl}. We use the 3-layer GIN with 32 hidden unit dimensions and sum readout function. We utilize the LIBSVM with default setting in sklearn \cite{sklearn}\footnote{scikit-learn.org, BSD-3-Clause license} as the downstream classifier.  
\textbf{Unsupervised node classification.} 
Following \cite{dgi} we employ 2-layer GCN as the encoder with 512 hidden units and  $l_2$-regularized logistic regression classifier in sklearn \cite{sklearn} for downstream tasks.  
\textbf{Graph classification in transfer learning} 
We pre-train the encoder on ZINC dataset and fine-tune on 4 other datasets. We utilize the 5-layers GIN encoder with 300 hidden units to keep consistency with \cite{pretrain}. A linear classifier is added on top of graph-level embeddings to predict downstream graph labels. In fine-tuning process, parameters of the pre-trained encoder and the linear classifier are end-to-end optimized together. We use the data processed in pytorch-geometric form provided by \cite{pretrain}.

% Both downstream classier are trained with 10-fold cross validation. 
All datasets for unsupervised graph and node classification tasks can be downloaded and pre-processed with pytorch\_geometric. For  all three pretext tasks, we train the model with Adam optimizer with learning rate $0.001$, epochs 500, and  batch size $256$ following previous works. We also utilize the early stop mechanisms. The split for train/validation/test sets for all downstream tasks is 80\%:10\%:10\%. 

\section{Comparison of Augmentation Methods}
\label{app:compare_augmentation}
\begin{figure}[tb!]
\begin{minipage}[b]{1\textwidth}
\subfigure[COLLAB]{\includegraphics[width=0.24\textwidth]{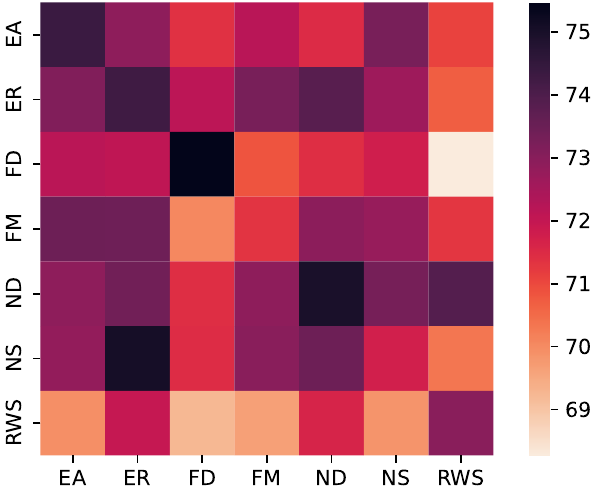}}
\subfigure[DD]{\includegraphics[width=0.24\textwidth]{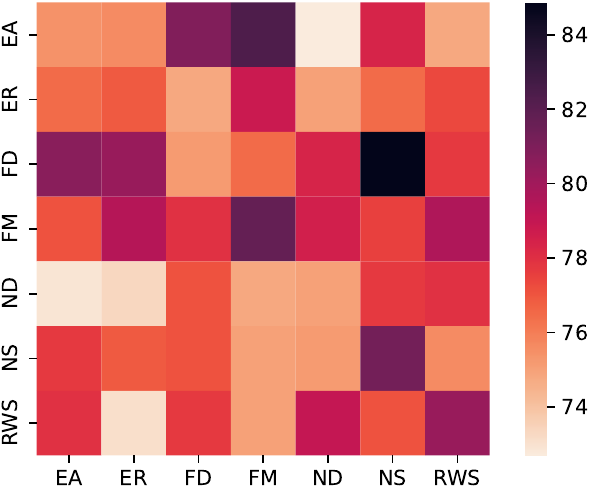}}
\subfigure[IMDB-B]{\includegraphics[width=0.24\linewidth]{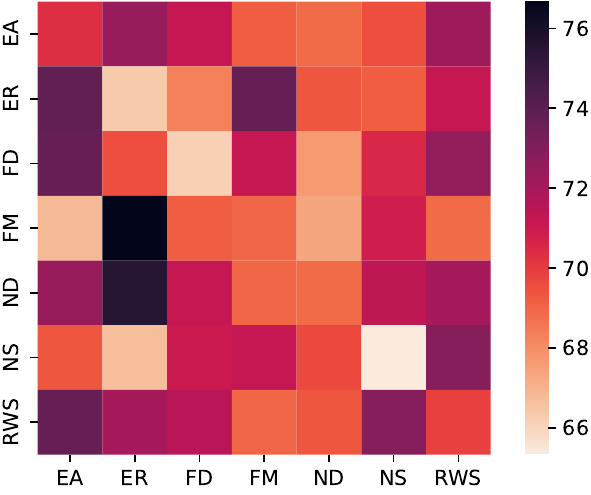}}
\subfigure[IMDB-M]{\includegraphics[width=0.24\linewidth]{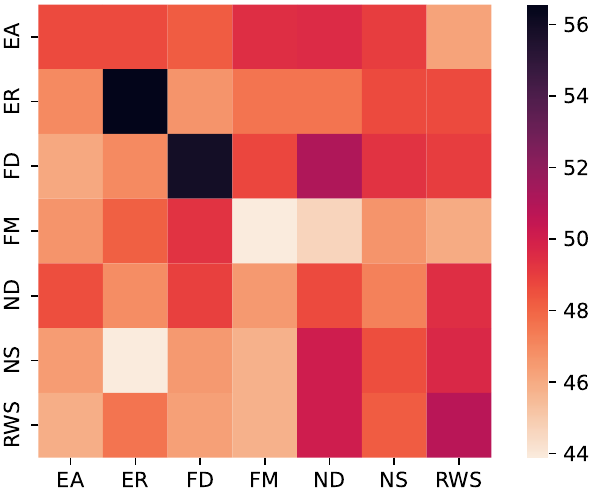}}
\end{minipage}
\begin{minipage}{\linewidth}
\subfigure[NCI1]{\includegraphics[width=0.24\linewidth]{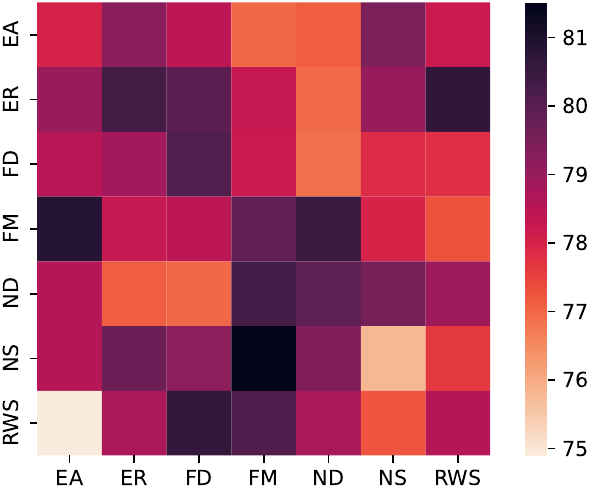}}
\subfigure[PROTEINS]{\includegraphics[width=0.24\linewidth]{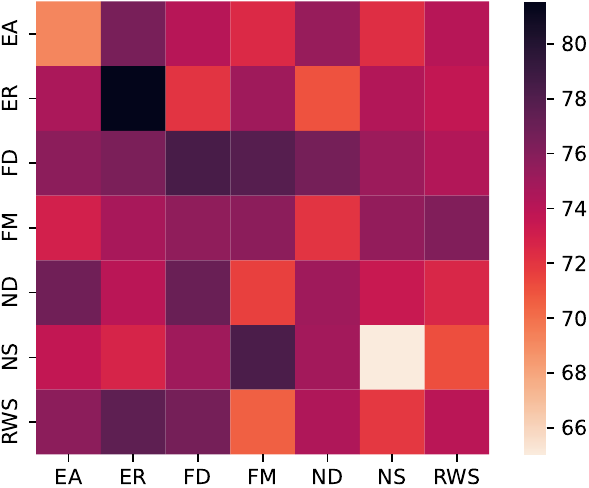}}
\subfigure[RDT-B]{\includegraphics[width=0.24\linewidth]{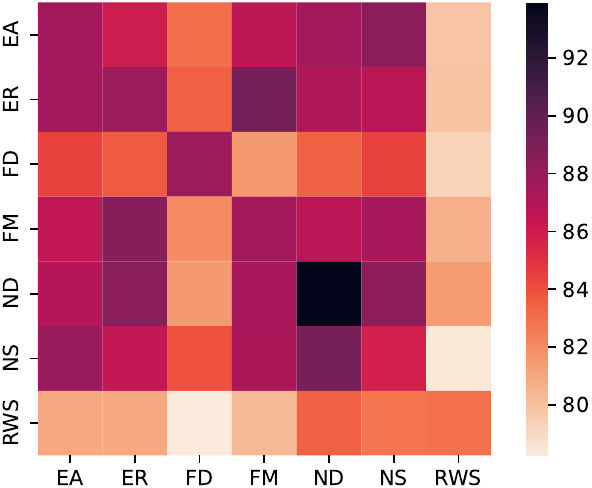}}
\subfigure[RDT-M]{\includegraphics[width=0.24\linewidth]{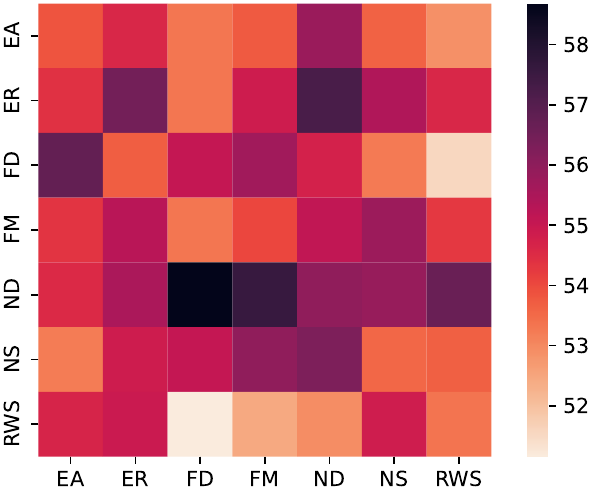}}
\end{minipage}
\vspace{-5pt}
\caption{Comparison of augmentation methods on 8 datasets.}
\vspace{-10pt}
\label{fig:comp_aug}
\end{figure}
Since our methods is asymmetrical, we conduct experiments on using different augmentations in settings of unsupervised graph classification tasks. 
All the settings and configurations remain consistent with previous unsupervised graph classification tasks. 
We exhaustively compare the every combination of two graph augmentation methods including Edge Add (EA), Edge Remove (ER), Feature Dropout (FD), Feature Masking (FM), Node Drop (ND), Node Shuffling (NS), Random Walk Subgraph (RWS) on 8 datasets. 
The augmentation hyper-parameter $p$ is set to 0.7 for all combinations. 
The results of 5-trials' mean are shown in Fig.~\ref{fig:comp_aug}, from which we make the following observations. 
First, the performance varies significantly when using different augmentation approaches. For example, for COLLAB dataset, the combination of FD and RWS only achieves 68.27\% accuracy while the combination of NS and ER achieves 75.07\% accuracy. 
Thus, we argue that using compound augmentation approach, e.g., discard a portion of nodes and mask a portion of nodes at the same time for one view, could be a direction for further research on graph contrastive learning. 
% Adaptively or leanablely compounding the augmentation approaches may stabilize and improve the self-supervised performance. 
Self-supervised classification performance may be stabilized and improved by adaptively or learnablely compounding the augmentation approahces. 
Second,  we identify that on some datasets such as COLLAB, IMDB-B, NCI1, using two identical augmentation approaches for two augmented views obtain relatively good even the best classification results. 
% Recall the observation in Sec.~\ref{sec:exp_result}, a randomly initialized untrained encoder obtains a competitive result. 
This observation provides reference for how to design augmentation schemes for graph domain.

\end{document}